\documentclass[twoside]{article}
\usepackage{aistats2020}

\usepackage[ruled, vlined, linesnumbered]{algorithm2e}
\usepackage{tikz}
\usepackage{multirow, makecell}
\usepackage{booktabs}

\usepackage{amsmath, amsthm, amssymb, bm}
\usepackage{mathtools}

\usepackage{xcolor}
\usepackage{graphicx}
\usepackage{subfig}
\usepackage{floatrow}

\usepackage{graphicx} 
\usepackage{times}
\usepackage{helvet}
\usepackage{courier}
\usepackage{tabu}

\usepackage{hyperref}
\usepackage{caption}
\usepackage{float}
\usepackage{multirow}
\usepackage{makecell}

\begin{document}

\twocolumn[
\aistatstitle{Scalable Hybrid HMM with Gaussian Process Emission \\ 
for Sequential Time-series Data Clustering}

% \aistatsauthor{ Yohan Jung \And JinkYoo Park }
% \aistatsaddress{ KAIST & Industrial Systems \& Engineering \And KAIST & Industrial Systems \& Engineering }

\aistatsauthor{ Yohan Jung \And Jinkyoo Park }
\aistatsaddress{ KAIST  \And KAIST  }
% \aistatsauthor{  }
% \aistatsaddress{  }
]

\begin{abstract}
Hidden Markov Model (HMM) combined with Gaussian Process (GP) emission can be effectively used to estimate the hidden state with a sequence of complex input-output relational observations. Especially when the spectral mixture (SM) kernel is used for GP emission, we call this model as a hybrid HMM-GPSM. This model can effectively model the sequence of time-series data. However, because of a large number of parameters for the SM kernel, this model can not effectively be trained with a large volume of data having (1) long sequence for state transition and 2) a large number of time-series dataset in each sequence. This paper proposes a scalable learning method for HMM-GPSM. To effectively train the model with a long sequence, the proposed method employs a Stochastic Variational Inference (SVI) approach. Also, to effectively process a large number of data point each time-series data, we approximate the SM kernel using Reparametrized Random Fourier Feature (R-RFF). The combination of these two techniques significantly reduces the training time. We validate the proposed learning method in terms of its hidden-sate estimation accuracy and computation time using large-scale synthetic and real data sets with missing values.
\end{abstract}

% The combined learning manners enables to train massive sequential time-series dataset.

\section{Introduction}

% Sequential time-series data describes the evolution of the target system over time. Recognizing the system's change by this data has been one of the important research topic. 

% \textcolor{blue}{a sequence of vibration time-series can represent the state of various mechanical structures}. 

A sequence of time-series data is ubiquitous. Such a sequence of time-series data can often describe the evolution of the hidden state of a target system. For example, a sequence of blood pressure or heartbeat time-series can represent the health condition of a patient. Therefore, inferring the hidden state of a target system using a sequence of time-series has long been an important research topic in engineering and science.

HMM is a model designed to specify the varying hidden states for sequential data. HMM has been mainly applied to classifying human action, speech, and DNA sequence \cite{gales2008application,yamato1992recognizing,ahmad2006hmm,yang1997human,haussler1996generalized}, which are all related to understanding the evolution of target system characteristics over time. In general, HMM consists of two-part; The transition model considers Markov's dynamic of discrete hidden state, and the emission model considers the likelihood of the observed data in a given hidden state. Since the emission model of vanilla HMM has a limitation in expressing complex observations because of its simple assumption, Hybrid HMM whose emission considers a nonlinear function such as neural net (NN), has been devised \cite{li2013hybrid,le2013emotion,yang2015deep,krishnan2016structured,tran2016unsupervised}. However, even though Hybrid HMM shows better performance with improved modeling power, it still has limitations in applying to incomplete data with missing values \cite{uvarov2019imputation}.

To address this problem, HMM using GP as the emission has been considered because GP has excellent expressive power as well as a nonparametric characteristic of GP enables the combined model applicable to the data containing missing values. As examples of these attempts, Nakamura uses Semi-HMM with GP emission to classify dynamic motion \cite{nakamura2017segmenting}. Zitao also conducts a study that leads to improved clinical time-series data prediction by combining GP with the emission of the Kalman filter, a continuous hidden state version of HMM \cite{liu2012state}. Regarding time-series data, Hensman proposes a hierarchical model that combines the Dirichlet process with a GP mixture model, which clustered gene expression time-series data with missing values \cite{hensman2013hierarchical}. However, these studies consider the commonly used RBF kernel, the Periodic kernel, and the combination of them. Since choosing a kernel is what determines the covariance structure of the GP that describes the data, choosing the kernel should be considered more carefully.

% Subsequent study also suggests a fast learning method of the proposed model \cite{hensman2014fast}.

One way to solve the mentioned issue is to employ a Spectral Mixture (SM) kernel that can approximate any stationary kernel \cite{wilson2013gaussian}. It has also shown excellent performance when modeling time series data sets. Ulrich et al. \cite{ulrich2014analysis,ulrich2015gp} propose an Infinite HMM whose emission consists of the Mixture of Experts model using Multi-output GP with each output being modeled by a spectral mixture kernel. Then, the proposed model is used to analyze human brain signals that change over time. Their approach classifies human brain signals over time into distinct hidden states, which could be described by different covariances characterized in the spectral domain. However, this model can not be scalably trained with large scale time-series dataset because many parameters of SM kernel tend to require extensive training time and induce over-fitting.

In this paper, we first introduce the hybrid HMM with GP Emission using spectral mixture kernel, which is a simplified version of Ulrich's model for sequential single output time-series data. Then, we propose a scalable learning method of the introduced model for large sequential time series data. Specifically, we employ Stochastic Variational Inference (SVI) based on \cite{hoffman2013stochastic,johnson2014stochastic,foti2014stochastic} to efficiently process the long sequence of state-transition if long sequence dataset is given. Also, we develop Spectral Kernel Approximation by Reparameterized Random Fourier Feature (R-RFF) to efficiently process a large number of observations ($\geq 1000$) observed at each state. This approximation can reduce computational cost to optimize kernel hyperparameters for GP emission. As a result, the combination of these two methods enables our introduced model to train a large volume of the dataset.

Also, we validate the classification performance of the model trained by our proposed method on large-scale data containing missing values.

\raggedbottom

\section{Background}
We first discuss the model structures for HMM and GP. In addition, we discuss the background on the SM kernel and Random Fourier Feature (RFF) technique for approximating the SM kernel.

\subsection{Hidden Markov Model (HMM)}
HMM is a probabilistic model to depict the sequential observations obtained under the varying hidden state of a target system. HMM consists of a transition model describing the transition of the hidden state over time and an emission model depicting the probability distribution of the observations given the hidden state. For the transition of the hidden state, the Markov assumption is used.

Let $ Y = \{y_t\}_{t=1}^{T}$ be observations measured at each $t$ time and $ Z= \{z_t\}_{t=0}^{T}$ be the corresponding the discrete hidden states where $z_t \in \{1,..,K\}$. Then, the joint distribution of hidden state and observation of HMM is expressed as : 
\begin{align}
& p(Z,Y) = {\pi}_0(z_{0})\prod_{t=1}^{T} p(z_{t}|z_{t-1},A) p(y_{t}|z_{t}) 
\end{align}
where $A$ is the transition matrix of hidden states that $A_{i,j}$ represents the probability of the transition from $z_{t-1} = i$ to $z_t = j$, i.e $Pr(z_{t}=j|z_{t-1}=i) = A_{i,j}$ . The ${\theta}$ denotes the parameter of emission. The initial parameter ${\pi}_0$ plays role in setting where the hidden state occurs at the first time. For the emission $p(y_{t}|z_{t})$, Gaussian or Mixture of Gaussian distribution is usually used in case of continuous observations. To relax the over-fitting problem in training HMM, Bayesian HMM considers the parameters $\{ {\pi},A,{\theta} \}$ as random variable and aims to infer these variables using observations. Typically, Dirichlet distribution ($\mathrm{Dir}$) and Normal-inverse-Wishart distribution ($\mathrm{NIW}$) are assumed as the prior distribution for the parameters $\pi,A$ and $\theta$, respectively.

To train HMM, the Expectation Maximization (EM) is applied, which alternatively optimizes the parameters of distributions for the hidden state and the emission model. This alternation respectively corresponds to updating local variables and global variables. For the Bayesian HMM, Variational Bayes EM (VBEM) and Markov Chain Monte Carlo (MCMC) are used for training \cite{beal2003variational,xie2002learning}.

% GP regression is a nonparametric regression using GP as prior distribution on target variable . 
\subsection{Gaussian Process (GP)}
GP is a stochastic process which assumes that any finite random variables of the stochastic process follow Gaussian distribution. GP can be used as a prior for a function that explains the relation between input and output \cite{rasmussen2003gaussian}. Let $x$ and $y$ be the pair of inputs and outputs and let $f$ be target function with GP prior assumption to consider the relation between $x$ and $y$.
\begin{align}
& y = f(x) + \epsilon, \  \epsilon \sim N(0,{{\sigma}_{\epsilon}}^2) \\
& f(x) \sim GP \left( m(x) ,k(x,x';\theta)  \right)
\end{align}
where $m(x)$ denotes the mean function of $f$. The kernel function $k(x,x';\theta)$ defines $\mathrm{cov}(f(x),f(x'))$ by the inputs $x,x'$ and the hyperparameter $\theta$.

The kernel hyperparameter $\theta$ is trained by maximizing the log marginal likelihood $\log{p(Y|X)}$ for the given inputs $X = \{ x_{n} \}_{n=1}^{N}$ and outputs $Y = \{ y_{n} \}_{n=1}^{N}$ with the number of dataset $N$.
\begin{align}
 \log{p(Y|X)} &= -\frac{1}{2} {Y}^T(K_{X,X} +{{\sigma}_{\epsilon}}^{2}I)^{-1}Y  \nonumber  \\
& -\frac{1}{2}\mathrm{log} |K_{X,X} +{{\sigma}_{\epsilon}}^{2}I| 
-\frac{n}{2} \log{2\pi} 
\end{align}
where $K_{X,X}$ is the evaluation of kernel function over $X$, i.e. $[K_{X,X}]_{ij} = k(x_{i},x_{j})$.  

% For prediction on ${X}^{*}$, the posterior predictive distribution of $f({X}^{*})$ can be computed by 
% \begin{align}
%  & p(f^{*}|X^{*},X,Y) = N \big(f^{*}|{\mu}^*,{\Sigma}^*) \nonumber \\
%  & {\mu}^* = K_{X^{*},X}(K_{X,X} +{{\sigma}_{n}}^{2}I)^{-1}Y \\
%  & {\Sigma}^* = K_{X^{*},X^{*}} - K_{X^{*},X}(K_{X,X}+{{\sigma}_{n}}^{2}I)^{-1}K_{X,X^{*}} 
% \end{align}

\subsection{Spectral Mixture Kernel (SM)}
\noindent For the stationary kernel which assumes that covariance between $f(x_{1})$ and $f(x_{2})$ for any inputs $x_{1},x_{2} \in \mathrm{R}^{P}$ is invariant to translation of the inputs, i.e. $cov(f(x_{1}),f(x_{2})) = k(|x_{1}-x_{2}|,0)$ \cite{bochner1959lectures}. 
%  Bochner's theorem implies the foundation to find the structure of the stationary kernel automatically

By Bochner's theorem, stationary kernel $k(\tau)$ can be obtained by taking the Fourier transform to spectral density of $S(s)$ for target function $f$ where ${\tau} = |x_{1}-x_{2}|$ between two inputs $x_{1}$ and $x_{2}$.
\begin{align}
& k(\tau) = \int e^{-2\pi i s^{\mathrm{T}} {\tau} } S(s)  ds 
\end{align}
This theorem implies that if $S(s)$ approximates well the empirical spectral density of observations $\{y_{t}\}_{t=1}^{T}$ of $f$, the corresponding kernel of target function $f$ can be obtained by applying Fourier transform to the $S(s)$. Wilson \cite{wilson2013gaussian} implements this idea to generate a new kernel known as the spectral mixture kernel (SM kernel). 

This approach first represents the spectral density $S(s) = \frac{1}{2}\left( \psi(s) + \psi(-s) \right) $ using the weighted sum of Gaussian distribution $\psi(s)$ to define the induced $k(\tau)$ on a real domain.
\begin{align}
\psi(s) &= \sum_{q=1}^{Q} w_{q} N(s|\mu_{q},\Sigma_{q}) 
\end{align}
where $\mu_{q} = ( \mu_{q}^{(1)},..,\mu_{q}^{(P)} )$ and $\Sigma_{q} = \mathrm{diag}( \nu_{q}^{(1)},..,\nu_{q}^{(P)} )$.
Putting $S(s)$ into the equation $(5)$ leads the SM kernel defined as
\begin{align}
& k_{SM}(\tau) = \sum_{q=1}^{Q} w_{q} \mathrm{cos} \left(2\pi {\tau}^T {\mu_{q}} \right) \prod_{p=1}^{P} \mathrm{exp} \left( -2{\pi}^2{\tau}_{p}^2{\nu_{q}^{(p)}} \right) 
\end{align}
\noindent where ${\tau}_{p}$ is the $p$ th components in the ${\tau} \in \mathrm{R}^{P}$.

\subsection{Sparse Spectrum GP Approximation by Random Fourier Feature (RFF) }
\subsubsection{Random Fourier Feature (RFF)}
Random Fourier Features method has been proposed to reduce the computation cost in the kernel learning domain \cite{rahimi2008random}. This method approximates the kernel function $k(x-y)$ by applying the Monte Carlo integration \cite{caflisch1998monte} to the Bochner's theorem $(5)$.

Let $S =\{ s_{i}\}_{i=1}^{M}$ be the sampled $M$ spectral points from the spectral density $S(s)$. Then, the approximated kernel $\hat{k} (x-y)$ is obtained by the equation : 
\begin{align}
 k(x-y) &= \mathbb{E}_{s \sim S(s)} [ e^{2\pi i s^{\mathrm{T}}(x-y) } ] \nonumber \\
& \approx \frac{1}{M} \sum_{i=1}^{M} \cos({2\pi s_i}^{T}x) \cos({2\pi s_i}^{T}y) \nonumber \\
 & \qquad\qquad\qquad + \sin({2\pi s_i}^{T}x)\sin({2\pi s_i}^{T}y)
\end{align}

This can also be defined by the inner product of feature map. Let $ x_{S} = [2 \pi x^{T} s_{1},.., 2 \pi x^{T} s_{M}] $ be the feature inputs and $ \phi_{S}(x) = \frac{1}{\sqrt M } [\cos( x_S ), \sin(x_S )] \in R^{2M} $ be the feature map. Then the approximated kernel $\hat{k} (x-y)$ can also expressed as $\phi_{S}(x)\phi_{S}(y)^{T}$ based on $(8)$. Then, the corresponding Gram matrix $K_{X,X}$ can be computed as $\Phi_{S}(X)\Phi_{S}(X)^{T}$ where $\Phi_{S}(X) = [\phi_{S}(x_1);..;\phi_{S}(x_n)] \in R^{n \times 2M}$.

\subsubsection{Sparse Spectrum Approximation in Gaussian Process}
Approximated kernel by RFF can reduce the training and inference time compared with original GP \cite{quia2010sparse}. 

When the approximate kernel by $(8)$ replace $K_{X,X}$, then the log marginal likelihood $(4)$ is differently computed with the sampled spectral points $S =\{ s_{i}\}_{i=1}^{M}$.
\begin{align}
\log{p(Y|X,S)} &= -\frac{1}{2} {Y}^T(\Phi_{S}(X)\Phi_{S}(X)^{T} +{{\sigma}_{\epsilon}}^{2}I)^{-1}Y \nonumber \\ 
& -\frac{1}{2}\mathrm{log}|\Phi_{S}(X)\Phi_{S}(X)^{T} +{{\sigma}_{\epsilon}}^{2}I|  -\frac{n}{2} \log{2\pi} 
\end{align}
Since the computation of $(\Phi_{S}(X)\Phi_{S}(X)^{T} +{{\sigma}_{\epsilon}}^{2}I)^{-1}$ takes 
$O(nM^2)$ instead of $O(n^3)$ by matrix inversion lemma \cite{watkins2004fundamentals}. We can reduce the computation time when $M \ll n $. 

% The corresponding predictive distribution is also similarly computed. 
% \begin{align}
% & p(f^{*}|X^{*},X,Y,\Phi_{S})) = N \big(f^{*}|{\mu_{ssgp}}^*,{\Sigma_{ssgp}}^*) \nonumber \\
% & {\mu_{ssgp}}^{*} = \Phi_{S}(X^{*})A^{-1}\Phi_{S}(X)^{T}Y \\
% % & {\Sigma_{ssgp}}^{*} = {{\sigma}_{\epsilon} }^{2} + {{\sigma}_{\epsilon} }^{2}\Phi(X^{*})A^{-1}\Phi(X^{*})^{T}
% & \color{blue}{
% {\Sigma_{ssgp}}^{*} = {{\sigma}_{\epsilon} }^{2}\Phi_{S}(X^{*})A^{-1}\Phi_{S}(X^{*})^{T} }
% \end{align}
% where $A = \Phi(X)_{S}^{T}\Phi_{S}(X) + ({\frac{ \sqrt{2M}{\sigma}_{\epsilon} }{{\sigma}_{0}}})^{2} I_{2M} $.

\section{Methodology}

\subsection{Hybrid HMM with GP Emission}

\begin{figure}[h]
\setcounter{figure}{1}
\centering
% {\includegraphics[width=.99\linewidth,height = 4cm]{EXPNEW1_v1/hybrid_hmmgp_graphic_v3.PNG}} 
{\includegraphics[width=.99\linewidth]{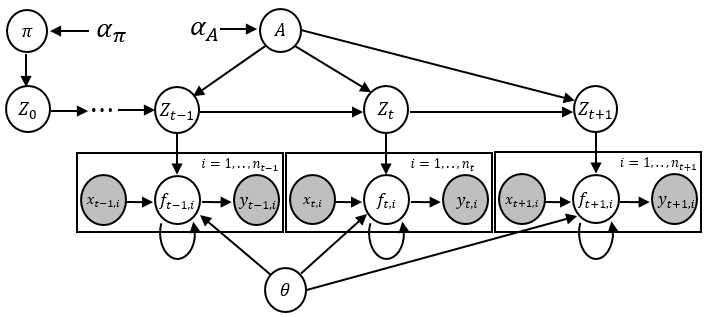}} 
\caption{Hybrid HMM with GP Emission}
\end{figure}

% We consider that $n_t$ time-series inputs $\{x_{t,j}\}_{j=1}^{n_t}$ and outputs $\{y_{t,j}\}_{j=1}^{n_t}$ every time step $t$ is obtained sequentially where $y_{t,j}$ is the $j$ th observation given at time $t$ and $x_{t,j}$ be the corresponding input. The $\{ y_{t,:},x_{t,:} \}$ and $z_t$ denote the full observations, corresponding inputs given at time $t$ and the corresponding hidden state over $\{y_{t,:},x_{t,:}\}$ respectively. 

\noindent We introduce a hybrid HMM with GP emission using SM kernel designed to classify the hidden state for a sequence of single-output time-series data. 

We assume that the output time-series data $\{y_{t,j}\}_{j=1}^{n_t}$ corresponding to time-series inputs $\{x_{t,j}\}_{j=1}^{n_t}$ are obtained every time step $t$. Here, $x_{t,j}$ and $y_{t,j}$ are the $j$ th input and output given at time $t$, respectively. $\{ y_{t,:},x_{t,:} \}$ denote the full observations at time $t$. $z_t$ denotes the corresponding discrete hidden state over $\{y_{t,:},x_{t,:}\}$ where $z_t \in \{1,..,K\}$.

Hybrid HMM with GP emission follows the same structure of vanilla HMM with equation (1) except the emission function. The emission of this model introduces the GP prior on the $f_{t,:}$ to explain the relation between $y_{t,:}$ and $x_{t,:}$ under the hidden state $z_t$ and SM kernel $k_{z_t}$ as
\begin{align}
f_{t,:}|x_{t,:},z_{t} &\sim GP \left( m_{z_{t}}(x_{t,:}) ,k_{z_{t}}(x_{t,:} , x_{t,{:}};\theta_{z_{t}} )  \right) 
\end{align}
where $\theta_{z_{t}}$ denotes the hyperparameters of the SM kernel $k_{z_{t}}$ having $Q_{z_{t}}$ mixture components. In addition, each point in time series is measured with independent and identically distributed (i.i.d.) Gaussian noise as
\begin{align}
y_{t,i}  &= f_{t,i}(x_{t,i}) + \epsilon , \  \epsilon \sim N(0,{{\sigma}_{\epsilon}}^2) 
\end{align}
Then, the log likelihood of the observation $p(y_{t,:}|x_{t,:},z_{t})$ is computed as the marginalization over the prior $f_{t,:}$ given $k_{z_{t}}$ as:
\begin{align}
 p(y_{t,:}|x_{t,:},z_{t}) &= \int p(y_{t,:}|f_{t,:},x_{t,:},z_{t})p(f_{t,:}|x_{t,:},z_{t}) df_{t,:} 
\end{align}
Additionally, we assume the prior distribution for transition matrix $A$ and initial distribution $\pi$ where the Dirichlet distribution is assigned to each row of $A$ and $\pi$. 
\begin{align}
& p({\pi}) = \mathrm{Dir}({\pi}|{\alpha}^{\pi}) \\
& p(A) = \prod_{i=1}^{K} \mathrm{Dir}(A_{i,:}|{\alpha}_{i}^{A}) 
\end{align}

\subsection{Variational Bayes Expected Maximisation (VBEM)}

\noindent HMM with GP emission can be trained by Variational Bayes Expected Maximisation (VBEM). Given the model assumptions for HMM with GP, the log joint likelihood of $Z,Y$ given $X$ is expressed as
\begin{align}
\log{p(Z,Y|X)} &= \log{{\pi}_0(z_{0})} + \sum_{t=1}^{T} \log {p(z_{t}|z_{t-1},A)} \nonumber \\
& + \sum_{t=1}^{T} \log{p(y_{t,:}|z_{t},x_{t,:})} 
\end{align}

In VBEM approach, variational distribution $q(\pi,A,Z)$ are introduced to approximate the posterior $p(\pi,A,Z|Y,X)$. Mean field assumption for $q(\pi,A,Z)$ leads to independent relation for each variable, i.e $q(\pi,A,Z) = q(\pi)q(A)q(Z)$. After we apply the Jensen's inequality to the log marginal likelihood $\int {\log{p(Z,Y|X)}p(\pi)p(A)} d\pi dA dZ$, we get the following Evidence Lower Bound $\mathcal{L}$ (ELBO) :
\begin{align}
& \mathcal{L}(q(\pi),q(A),q(Z)) = \mathbb{E}_{q}[\log{p(Z,Y|\pi,A)} - \log{q(Z)}] \nonumber   \\ 
& + \mathbb{E}_{q}[\log{p(\pi)} - \log{q(\pi)}]  + 
\mathbb{E}_{q}[\log{p(A)} - \log{q(A)}]
\end{align}

Maximizing the $\mathcal{L}$ is equivalent to minimize the KL divergence between $q(\pi,A,Z)$ and $p(\pi,A,Z|Y,X)$, which results in updating the variational parameters $\{w_{\pi},w_{A} \}$ of global variable $\{\pi,A 
\}$ and local variable $Z$ alternatively.

The update of local variable $q(Z)$ is induced as follows: 
\begin{align}
 q^{*}(Z) & \propto \exp \Big( \sum_{t=1}^{T} \log{ p(y_{t,:}|z_{t},x_{t,:}) }  \nonumber \\ 
&+ \mathbb{E}_{q(\pi)} [\log{p(z_{0})}] 
 + \sum_{t=1}^{T} \mathbb{E}_{q(A)}[ \log{p(A_{z_{t-1},z_{t}}) } ]   \Big) 
\end{align}
To evaluate the updated local variable $q^{*}(Z)$, the following auxiliary variables $\widetilde{\pi},{\widetilde{A}_{j,i}}$ should be computed.
\begin{align}
& \widetilde{\pi} = \exp{(\mathbb{E}_{q(\pi)} [ \log{\pi_{j}} ])}  \  ,  \  {\widetilde{A}_{j,i}} = \exp{( \mathbb{E}_{q(A)} [ \log{A_{j,i}} ])}
\end{align}
Then, forward-backward algorithm using $p(y_{t,:}|z_{t},x_{t,:})$, $\widetilde{\pi}$ and $\widetilde{A}$ generates $q^{*}(z_{i}),q^{*}(z_{i-1},z_{i})$ to be used for the update of global variable parameters.

Updating the global variable parameters is conducted as follows:
\begin{align}
& w_{\pi_{j}} =  {\alpha}_{j}^{\pi} + \mathbb{E}_{q(Z)} [1_{z_{0}=j}] \\
& w_{A_{j,i}} = {\alpha}_{j,i}^{A} +  \sum_{t=1}^{T} \mathbb{E}_{q(Z)} [ 1_{z_{t-1}=j,z_{t}=i} ]
\end{align}
where $w_{\pi_{j}}$ and $w_{A_{j,i}}$ denote variational parameter for j-th element of $\pi$ and $(j,i)$ element of $A$. The ${\alpha}_{j}^{\pi}$ and ${\alpha}_{j,i}^{A}$ denote Dirichlet prior for $\pi_{j}$ and Dirichlet prior for $A_{j,i}$. The $1_{z_{0}=j},1_{z_{t-1}=j,z_{t}=i}$ denote the sufficient statistic for $q(\pi)$ and $q(A)$, respectively. 

The hyperparameters of SM kernel $\theta = \{\theta_{1},..,\theta_{K}\}$ are trained by maximizing the following objective :
\begin{align}
&  \sum_{t=1}^{T} \mathbb{E}_{q(Z)} \left[ \log{p(y_{t,:}|z_{t},x_{t,:})} \right] \nonumber \\
&  = \sum_{t=1}^{T}\sum_{j=1}^{K} \log{p(y_{t,:}|z_{t}=j,x_{t,:})} q(z_{t}=j)  
\end{align}
This optimization applies the gradient method such as nonlinear conjugate gradients and L-BFGS.

\raggedbottom

\subsection{Scalable Learning}
The VBEM method is difficult to employ for training HMM-GP with the SM kernel using the data having 1) long sequence of state transition ($T$ is large), 2) a large number of observations in each time-series ($n_t$ is large), mainly due to a large number of the SM kernel parameters.

To tackle the mentioned issue, we apply Stochastic Variational Inference (SVI) for 1) based on \cite{johnson2014stochastic,foti2014stochastic}, which intend to approximate the ELBO for long sequence data by linear approximation through the sampled short sequence data. For 2), we develop Spectral Kernel Approximation by Reparameterized Random Fourier Feature (R-RFF) that optimizes SM kernel hyperparameters in a stochastic manner.

\subsubsection{SVI Approach}
In the SVI approach, we approximate the log likelihood of the total $T$ sequence of time-series data as the log likelihood of uniformly sampled short $L$ sequence of time-series data called by batch data with the single batch sampling. 

Given the $T$ sequence of time series observations, we randomly sample $L$ time series observations $Y_{L}^{s} = \{y_{i,:},...,y_{i+L-1,:} \}$ corresponding the inputs $X_{L}^{s}$ and the hidden states $Z_{L}^{s}$, where the index $i$ is sampled uniformly from $i \in \{1,...,T-L+1\}$.

Since the ELBO for sampled $\{ X_{L}^{s},Y_{L}^{s} \}$ needs to calibrate in order to replace the ELBO for $\{X,Y\}$, we take linear approximation of ELBO of $\{X,Y\}$ by considering expected log joint likelihood of $\{ Z_{L}^{s},Y_{L}^{s} \}$ given $X_{L}^{s}$.
\begin{align}
&  \mathbb{E}_{s} \Big[ \mathbb{E}_{q} [\log{p(Y_{L}^{s},Z_{L}^{s} | X_{L}^{s} )}] \Big] \nonumber  \\
& \approx \frac{1}{T-L+1} \mathbb{E}_{q} \Big[ \sum_{t=1}^{T-L+1} \log{p(z_{t-1})} +  \nonumber \\ 
& L \sum_{t=1}^{T} \log{p(A_{z_{t-1},z_{t}})} +  L \sum_{t=1}^{T} \log{p(y_{t,:}|z_{t},x_{t,:})} \Big] 
\end{align}
By the equation (22), the batch factor $c_{s}^{A},c_{s}^{\theta} $ to coordinate ELBO of $\{ X_{L}^{s},Y_{L}^{s} \}$ with full ELBO of $\{ X,Y \}$ is obtained as follows:
\begin{align}
&  c_{s}^{A} = \frac{T-L+1}{L} \ \ , \ \ c_{s}^{\theta} = \frac{T-L+1}{L} 
\end{align}
Applying stochastic natural gradient descent to the natural parameter of variational distribution $q(A),q(\pi)$ iteratively leads to the update rule for the global variable as follows :
\begin{align}
w_{\pi_{j}}^{n+1} &= (1-p_{n})w_{\pi_{j}}^{n} + p_{n}\left({\alpha}_{j}^{\pi} + \mathbb{E}_{q(Z^{s})} [1_{z_{i}=j}^{s}] \right)\\
\psi^{s} &=  c_{s}^{A} \ \mathbb{E}_{q(Z^{s})} \left[ \sum_{t=1}^{L} 1_{ z_{i+t-1}^{s} = j , z_{i+t}^{s} = k } \right]  \nonumber \\
w_{A_{j,k}}^{n+1} &= (1-p_{n})w_{A_{j,k}}^{n} + p_{n}({\alpha}_{j,k}^{A} + \psi^{s})
\end{align}
where $w_{\pi},w_{A}$ are the natural parameter of Dirichlet distribution and $p_{n}$ is $n$-th learning rate of stochastic optimization. SM kernel hyperparameters $\theta$ are trained by maximizing expected log marginal likelihood $\lambda^{s}$ with the batch factor $c^{\theta}_{s}$.
\begin{align}
\lambda^{s}_{i} & = c_{s}^{\theta} \ \mathbb{E}_{q(Z_{L}^{s})}
\left[ \sum_{t=0}^{L-1} \log{p(y_{i+t,:}|z_{i+t},x_{i+t,:})} \right] 
\end{align} 

For multiple $M$ batch data $\{Y^{s_m}_{L},X^{s_m}_{L},Z^{s_m}_{L}\}_{m=1}^{M}$, variational parameters are similarly updated as follows: 
\begin{align}
 \tau_{j}^{s_{m}} &=  \mathbb{E}_{q(Z^{s_{m}})} [1_{z_{m_{i}}=j}^{s_m}] \nonumber \\
w_{\pi_{j}}^{n+1} &= (1-p_{n})w_{\pi_{j}}^{n} + p_{n}  \left( {\alpha}_{j}^{\pi} + \frac{1}{M}\sum_{m=1}^{M} \tau_{j}^{s_{m}} \right)\\
w_{A_{j,k}}^{n+1} &= (1-p_{n})w_{A_{j,k}}^{n} + p_{n} \left( {\alpha}_{j,k}^{A} + \frac{1}{M}\sum_{m=1}^{M}\psi^{s_{m}} \right)
\end{align}

SM kernel hyperparameters are updated by maximizing the following objective: 
\begin{align}
& \frac{1}{M} \sum_{m=1}^{M} \lambda^{s_{m}}_{m_{i}}    
\end{align}
where $m_{i}$ is first sampled index of $m$ batch data.\\

Updating the parameters of the local variable is similarly updated as VBEM except that full dataset $\{ Y,X \}$ are replaced with sampled $\{ Y_{L}^{s},X_{L}^{s} \}$.

% This approximation can reduce significant computation cost for GP emission.

\subsubsection{SM Kernel Approximation for GP Emission}

To scale up the GP emission, we approximate the SM kernel by Reparameterized Random Fourier Feature (R-RFF). Then, the approximated SM kernel is employed to regularized sparse spectrum GP approximation for the reduction of computation cost. 

\noindent Given the parameters of SM kernel $\{w_{q},{\mu}_{q},\sigma_{q}\}_{q=1}^{Q}$, we sample spectral points from Gaussian distribution $N(S;{\mu}_{q},\sigma_{q})$ by reparametrization trick.
\begin{align}
& s_{q,i} = \mu_{q} + \sigma_{q} \circ \epsilon_{i} 
% \ \ \forall \ i=1,..,m_q
\end{align}
where $\epsilon_{i} \sim N(\epsilon;0,I)$ for $i=1,..,m_q$. The sampled spectral points $\textbf{\em s} = \cup_{q=1}^{Q} \{ s_{q,i} \}_{i=1}^{m_q} $ induces the feature map $\phi_{SM}(x)$, which can approximate $k_{SM}(x-y)$.
\begin{align}
&  \phi_{SM}(x) = \left[ \sqrt{w_{1}}\phi_{1}(x),..,\sqrt{w_{Q}}\phi_{Q}(x) \right] \nonumber \\
&  k_{SM}(x-y) \approx \phi_{SM}(x)\phi_{SM}(y)^{T}
\end{align}
where $\phi_{q}(x)$ is the feature map defined by the sampled spectral points $\{ s_{q,i} \}_{i=1}^{m_q}$ from $N(S;u_q,\sigma^{2}_{q})$ by (8).

% This lower bound can prevent each spectral distribution from collapsing to each other.
Based on Stochastic Gradient Variational Bayes (SGVB) \cite{kingma2013auto}, we derive the regularized lower bound of log marginal likelihood with Reparametrized kernel approximation (31).
\begin{align}
& \log{p(y_{t,:}|z_{t},x_{t,:})} \nonumber \\
&\geq \int \log{p(y_{t,:}|z_{t},x_{t,:},S)}q(S) d S - KL(q(S)||p(S)) \nonumber \\
& \approx \frac{1}{K}\sum_{k=1}^{K} \log {p(y_{t,:}|z_{t},x_{t,:},\textbf{\em s}^{(k)} )} - KL(q(S)||p(S))
\end{align}
where $\textbf{\em s}^{(k)}$ is repetitive $k$-th sampled spectral points from $q(S)=\prod_{q=1}^{Q} N(S;{\mu}_{q},\sigma_{q})$ for robust lower bound approximation. $p(S)$ is the prior distribution of spectral density, whose parameter can be tuned by using empirical spectral density. The second term $-KL(q(S)||p(S))$ of this lower bound (32) acts as a regularizer to prevent the spectral density distribution from collapsing in training.

The obtained (32) can be employed as the alternative to $\log{p(y_{t,:}|z_{t},x_{t,:})}$ with less computation.

\subsection{Computation Complexity} 

To analyze the computation complexity of our learning algorithm, we split the algorithm mainly into three parts; computation of log marginal likelihood for observations by GP emission, local update, and global update. We proceed with the analysis of our computation under a single batch assumption because the repetitive sampling for robust SVI and lower bound of log marginal likelihood (32) increases the total computation linearly.

For computing the log marginal likelihood of $T \times n$ observations with $K$ hidden state, the conventional VBEM approach costs $\mathcal{O}(KTn^{3})$.
Our approach costs $\mathcal{O}(KL nm^{2} )$, where the length of the sampled sequence is $L$ sequence, and the total $m$ sampled spectral points are used for SM kernel approximation. This is because SVI approach reduces to $\mathcal{O}(L)$ from $\mathcal{O}(T)$ and SM kernel approximation reduces to $\mathcal{O}(nm^{2})$ from $\mathcal{O}(n^{3})$. For the update of local variables, VBEM and SVI take $\mathcal{O}(K^{2}T)$ and $\mathcal{O}(K^{2}L)$ for the forward-backward algorithm, respectively. Updating the global variable is dominated by updating kernel hyperparameters. Computing the derivative of log marginal likelihood for each parameter costs $\mathcal{O}(n^3)$ \cite{rasmussen2003gaussian}. Thus, in the case of spectral mixture kernel, VBEM costs $\mathcal{O}((3Q+1)n^{3}KT)$ where All spectral mixture kernels take $Q$ Gaussian mixture components. However, our scalable approach takes $\mathcal{O}((3Q+1)nm^{2}KL)$.

In summary, our scalable learning method scalably trains the large dataset when we control $m$ and $L$ such that $nm^{2} \ll n^{3}$ and $L \ll T$.

\raggedbottom

\begin{table*}[htbp]
\begin{center}
    \scriptsize
    \setlength\tabcolsep{9.0pt}
    \extrarowsep=1.02ex
    {\begin{tabular}{@{\extracolsep{2pt}} l cc cc  cc cc @{}}
    % \footnotesize
    % \setlength\tabcolsep{3pt}
    % \extrarowsep= .8ex
    % {\begin{tabular}{@{\extracolsep{2.5pt}} l cc cc  cc cc @{}}
    \Xhline{2\arrayrulewidth}
    & \multicolumn{2}{c}{\bfseries Accuracy(Q=3)} 
    & \multicolumn{2}{c}{\bfseries {\#}Cluster(Q=3)} 
    & \multicolumn{2}{c}{\bfseries Accuracy(Q=6)} 
    & \multicolumn{2}{c}{\bfseries {\#}Cluster(Q=6)} \\
    \cline{2-3}
    \cline{4-5}
    \cline{6-7}
    \cline{8-9}
    \bfseries Methods   & \multirow{2}{*}{\bfseries 200Hz}  
                        & \multirow{2}{*}{\bfseries 1000Hz}
                        & \multirow{2}{*}{\bfseries 200Hz}  
                        & \multirow{2}{*}{\bfseries 1000Hz}
                        & \multirow{2}{*}{\bfseries 200Hz}  
                        & \multirow{2}{*}{\bfseries 1000Hz}
                        & \multirow{2}{*}{\bfseries 200Hz}  
                        & \multirow{2}{*}{\bfseries 1000Hz} \\
    \bfseries{(\#T, \#L, \#B-K, \#SPPt, \#SPPt-K)} &&& &&&  \\
    % ($\#F-Seq, \#B-Seq, \#K_{B}, \#SpecPt, \#K_{SP}$) &&& &&&  \\   
    \Xhline{2\arrayrulewidth}
    KNNDTW
    & .83 &  .92 
    & 8 & 8
    & .62 &  .69 
    & 8 & 8 \\    
    \hline
    HMM-T
    &  .39(.04)          & .34(.02)   
    &   8       &   8
    &  .30(.02)    & .27(.03)   
    &   7.8(.4) & 7.8(.4) \\        
    HMM-S
    & .86  & .77 
    & 8 & 8
    & .62(.04) & .42(.09)  
    & 8 & 8 \\        
    BHMM-T
    &  .48        & .46   
    &   8   & 8
    &  .40        & .46   
    &   8   & 8 \\        
    BHMM-S
    & \textbf{.89}  & \textbf{1.00}
    & 8    &  8
    & .74  & .78  
    & 8 & 7 \\            
    \hline
    GPSM-VBEM (100,\textemdash,\textemdash,\textemdash,\textemdash)
    & .82(.06)        &   \textemdash
    &  6.2(.60)        &   \textemdash  
    & \textbf{.87}(.03)        &   \textemdash    
    &  5.6(.40)        &  \textemdash  \\  
    \hline
    GPSM-SVI (100,10,1,\textemdash,\textemdash)
    &  .80(.06)       &  .82(.06)
    &   6.3(1.0)      &   5.0(.47)     
    &   .84(.06)      &  .67(.06)  
    &   5.4(1.3)      &   4.0    \\                 
    GPSM-SVI (100,10,3,\textemdash,\textemdash)
    &  .80(.06)       &.82(.10)
    &  6.1(1.0)      & 5.2(1.0)  
    & .81(.06)        &  \textemdash  
    &  5.4(.92)          &  \textemdash   \\       
    \hline
    GPSM(RSS)-SVI (100,10,1,5,2)              
    &  .86(.04)       &   .90(.07)
    &  6.3(1.0)        &   6.7(1.0)
    &  .78(.07)        &   .77(.05)
    &  5.0(.77)          &   5.3(.64)\\                 
    GPSM(RSS)-SVI (100,10,3,5,2)              
    & .86(.08)       &   .92(.07)   
    &  6.2(1.5)        &   6.3(1.4)
    & .84(.06)       &   .76(.04)
    &  6.4(.92)        &    5.4(.66) \\                 
    GPSM(RSS)-SVI (100,10,3,10,2)              
    & .87(.07)       &    .91(.07)  
    & 5.8(.98)        &     5.9(.30)
    & .80(.06)       &    .85(.05)
    & 5.8(.40)        &     5.7(.64) \\  
    GPSM(RSS)-SVI (100,10,3,10,4)              
    & .87(.06)       &      .90(.07)  
    & 5.9(.94)       &      5.9(.54)
    & .82(.06)       &      \textbf{.87}(.07)
    & 6.0(1.1)       &     5.6(.49) \\      
    \hline
    \Xhline{2\arrayrulewidth}
    \end{tabular}}    
    \caption{We report the result of a scalable Learning experiment for Q1. $\#$T and $\#$L denote the length of full and batch sequence, respectively. $\#$SPPt denotes the number of sampled spectral points of each Gaussian distribution for SM kernel. This means that a total of $\#$SPPt $\times$ Q spectral points are sampled for approximated SM kernel as one hidden state. $\#$B-K and $\#$SPPt-K denote the number of the sampled batch.}
\end{center}
\end{table*}

\begin{figure*}[ht]
  \begin{minipage}[b]{.55\linewidth}
  \captionsetup{justification=centering}
    \centering
    \subfloat[\small{Training time $\&$ Accuracy  }]
    {\includegraphics[width=\textwidth,height = 4.9 cm]{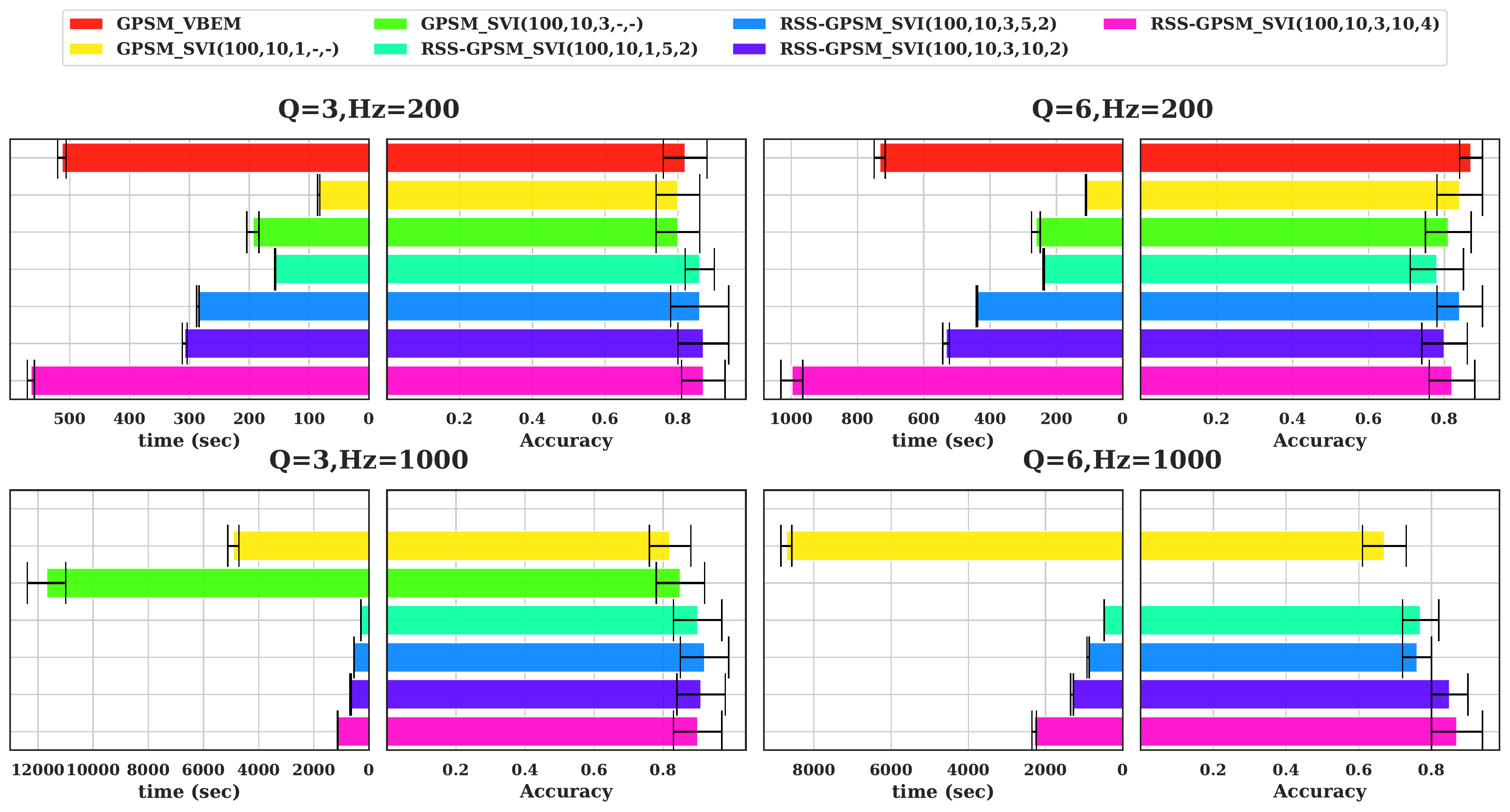}}
    % \subcaptionbox{\small{(a) GPSM(RSS)-SVI}}
  \end{minipage} \hfill
  \begin{minipage}[b]{.44\linewidth}
  \captionsetup{justification=centering}
    \centering{
    \subfloat[\small{Q=6, Hz=1000}]
    {\includegraphics[width=\textwidth,height=1.05cm]{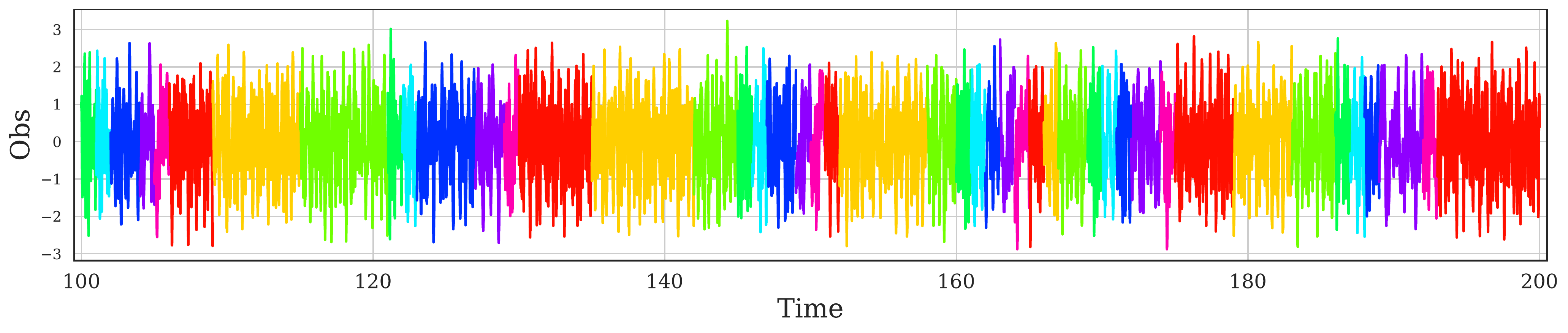}}
    \quad
    \subfloat[\small{GPSM(RSS)-SVI }]
    {\includegraphics[width=\textwidth,height=1.05cm]{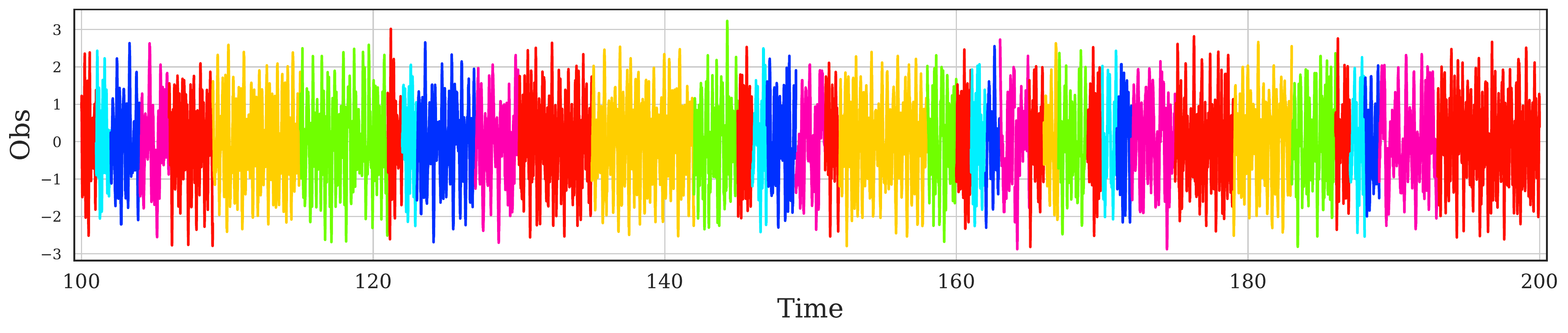}}
    \quad
    \subfloat[\small{KNNDTW}]
    {\includegraphics[width=\textwidth,height=1.05cm]{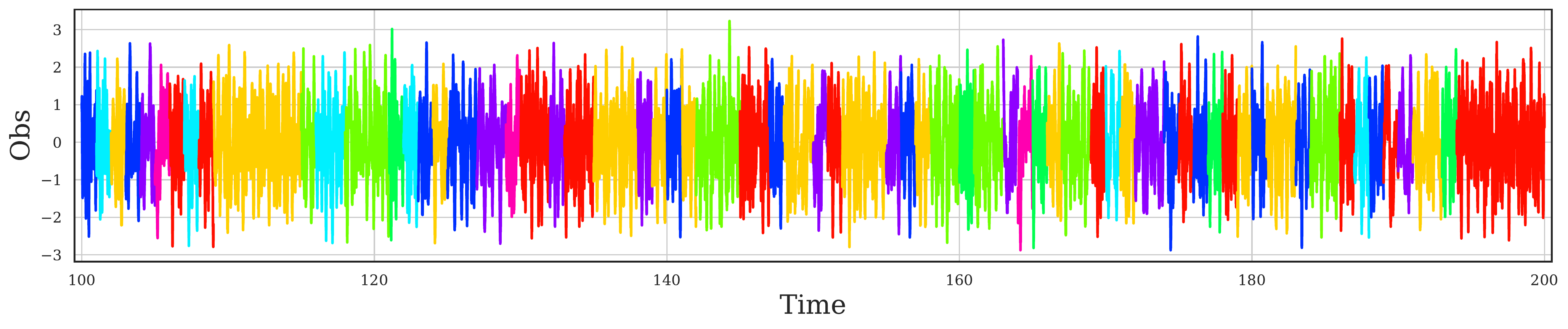}} 
  }
  \end{minipage}%
  \caption{(a) describes the results of the experiments for Hz=200 and Hz=1000 for Q1. This reveals the improvement by SVI and SM approximation through reparameterized RFF, respectively. Figures (b),(c), and (d) show the state Estimation Result for sinusoidal dataset Q=6,Hz=1000. (b),(c), and (d) correspond to the true coloring of testset and GPSM(RSS)-SVI and KNNDTW with accuracy .94 and .69, respectively.}
\end{figure*}

\section{Experiments}
We conduct two experiments for validation. In the experiments, the following performance metrics are used to evaluate the performance of the trained models.

\textbullet Accuracy : how correct the estimated hidden states match to the true label after reordering the estimated clusters by Munkres algorithm \cite{munkres1957algorithms}\\
% \textbullet $\log{p(Y|X)}$ : log marginal likelihood for the dataset \\
\textbullet Time : training time (seconds) for given dataset \\
\noindent \textbullet \#Cluster : the number of the estimated hidden states

\subsection{Scalable Learning (Q1)}

We conduct the first experiment to validate how the proposed learning strategies (SVI and R-RFF) expedite the training of HMM-GPSM with a large volume of data, i.e., 1) long sequence ($T$ is large) and 2) a large number of observations in each time-series ($n_t$ is large).

% whose different state is specified by the frequency component of observation and its amplitude. For state transition, we use Markovian assumption over each state. 

We use the synthetic sinusoidal dataset for the first experiment. We consider 8 states whose state transition follows a Markov assumption. Specifically, there are two groups of states whose state dynamics are different: $(1 \rightarrow 2 \rightarrow 3 \rightarrow 4)$ and $(5 \rightarrow 6 \rightarrow 7 \rightarrow 8)$. Each state except state 4 in the first group follows the stay probability $0.7$ and the move probability $0.3$. Similarly, each state except state 8 in the second group has stay probability $0.3$ and the move probability $0.7$. The special states $\{4,8\}$ connect the two group with probability 1, i.e., $(4 \rightarrow 5)$ and $(8\rightarrow 1)$.

Given a hidden state $s \in \{1,..,8\}$, we assume that the distinct sinusoidal $f_{Q}^{s}(t)$ is observed. 
\begin{align}
& f_{Q}^{s}(t) = \sum_{q=1}^{Q} \widetilde{\alpha}_{s,q}  \sin{ \left( \omega_{s,q} \pi \ t \right)} + \epsilon 
\end{align}
where $Q$ is the number of spectral components for sinusoidal. The frequency components $\omega_{s,q}$ is uniformly sampled from the interval $[0,20]$. The weight factors $\{ \widetilde{\alpha}_{s,1},..,\widetilde{\alpha}_{s,Q} \}$ are normalized version of $\{ {\alpha}_{s,1},.., {\alpha}_{s,Q} \}$ where ${\alpha}_{s,q}$ is uniformly sampled from the $[0,1]$. This normalization intends to make sure that the statistics such as the mean and standard deviation of each time-series observations are not be a determinant factor to classify the sinusoidal generated in different state.

To investigate the complexity of the spectral pattern, we consider $Q = 3,6$. In addition, to investigate the impact of the number of time-series measurements, we consider two sampling frequency 200Hz and 1000Hz. For every combination of $Q$ and sampling frequency Hz, we generate a time-series for $2T$ seconds using (33), i.e, $2T \times 200\mathrm{(Hz)}$ and $2T \times 1000\mathrm{(Hz)}$. Then, we use the first half of them for training ($T \times \mathrm{Hz}$) and the left half of them for test ($T \times \mathrm{Hz}$).

For the comparison of the proposed training methods, we take HMM-GPSM introduced in Figure 2 as the modeling baseline. Then, we compare mainly the following three learning strategies:

\textbullet GPSM-VBEM : Bayesian HMM with GP emission model combined with SM kernel trained by VBEM \\
\textbullet GPSM-SVI : Bayesian HMM with GP emission model combined with SM kernel trained by SVI \\ 
\textbullet GPSM(RSS)-SVI : Bayesian HMM with GP emission combined with approximated SM kernel trained by SVI  

To provide the information of how difficult the distinct sinusoidal data is classified, we also takes following benchmarks models based on HMM:

\textbullet HMM-T : Vanilla HMM with Multivariate Gaussian emission for time-series observation \\
\textbullet HMM-S : Vanilla HMM with Multivariate Gaussian emission model for spectral density observation by Discrete Fourier Transform \\
\textbullet BHMM-T : Bayesian HMM with Multivariate Gaussian emission for time-series observation \\
\textbullet BHMM-S : Bayesian HMM with Multivariate Gaussian emission model for spectral density observation by Discrete Fourier Transform 

We also consider a K-Nearest-Neighbors using Dynamic Time Warping (KNNDTW) which is commonly used for time-series clustering.

\textbullet KNNDTW : K-Nearest-Neighbors based on Dynamic Time Warping \cite{sakurai2007stream,hsu2011knn} \\
\raggedbottom

Table 1 reports the statistics of the mean and standard deviation for the accuracy and the number of estimated hidden states computed by 10 repetitive experiments. In terms of accuracy for the state estimation, the HMM-based on GPSM has comparable results with a vanilla version of HMM when $Q=3$ (data pattern is simple); however, the HMM-based on GPSM has the highest accuracy when $Q=6$ (data pattern is complex). Figure $3$, (a) shows clearly how the proposed training strategy (SVI+R-RFF) reduces the computational time without sacrificing accuracy. The reduction is especially significant when the number of data points in the time series is large. Figures $3$, (b), (c), and (d) shows the results of clustering for a different model, showing that the proposed method matches the true labels quite well.

% Yu et al. propose the revised forward-backward algorithm for HMM that allows the HMM and Semi-HMM to be trained with missing data \cite{yu2003hidden}.
\subsection{Robust State Estimation (Q2)}

In HMM research, handling the incomplete dataset containing missing values has been recognized as an important research topic because missing values induces the difficulty of training HMM and estimating the hidden state. To tackle this problem, YEH et al. have investigated how missing data affect the estimation of the HMM parameters in a clinical dataset where missing values are commonly found in the dataset \cite{yeh2012intermittent}. Popov et al. propose the algorithms of training HMM on sequence data set with missing values by marginalizing the likelihood of missing values for emission \cite{popov2016training}. Uvarov et al. develop the modified Viterbi algorithms of HMM for the imputation of incomplete motion data \cite{uvarov2019imputation}.

% Then, they compare the algorithm with standard method which fills out the missing values as the nearest values \cite{popov2016training}.

% This reveals the advantage of the GP emission as nonparametric model
In this experiment, we aim to evaluate how the HMM-GPSM trained by our proposed learning methods is robust with missing values in the large scale time-series dataset.

We use the PigArtPressure dataset from The UCR time series archive \cite{dau2018ucr}. This data set contains 312 Arterial-blood-pressure time-series data measured at different conditions. This dataset consists of 52 classes, and one time-series of each class has 2000 data points. For the experiment, we randomly chose the 10 class as the hidden state, and sample the observations from the bags of each class. To introduce state transition, we assign Markov assumption over each class. The probability of move and stay over each class is 0.5, respectively. This setting intends to reflect the interesting topic of recognizing the difference between stay and move for dynamic object \cite{anguita2012human,hallac2017toeplitz}. Then, we down-sample the timestamps into a half and then use the training set ($100 \times 1000$) and test set ($50 \times 1000$).

For comparison, we take GPSM(RSS)-SVI for our approach. Also, KNNDTW, HMM-S, BHMM-T, and BHMM-S which are used in previous experiment \textbf{4.1}, are compared again. Additionally, we consider the IOHMM \cite{bengio1995input,chiappa2003hmm} that has a different structure of HMM to specifically consider the relation between the inputs and outputs in emission:

\textbullet IOHMM : Input and Output HMM with Autoregressive Gaussian emission which assigns the inputs and outputs as spectral density observations and time-series observations\\ 

\raggedbottom

To make missing values in the dataset, we consider two different scenarios; 1) Randomly Missing (RM) that missing values are on randomly chosen timestamps, 2) Interval Missing (IM) that missing values consists of the intervals. We consider the two types of missing percentage $\{25,50\}$ over each 1000 time-series observations.

To process missing values for other benchmark models, we consider the conventional Fill-Out (FO) technique \cite{popov2016training} that recovers the missing value as the value of the nearest observation. In the case of our model, we also test the incomplete dataset without FO, which corresponds to NFO.

Table 2 compares the mean accuracy and the standard error for hidden state clustering, which are computed from 10 repetitive test experiments. For the RM case, all models achieve the clustering performance as good as the performance on the fully-observed dataset, implying that the FO technique is effective in filling missing observations for the RM case. However, for the IM case, the FO technique does not achieve good performance for simple HMM models and KNNDTW anymore; only our model can maintain the clustering accuracy both for IM25 and IM50. Note that Our method can even maintain the clustering performance even without applying FO techniques, which is impossible for simple HMM models and KNNDTW.

\begin{table}[t]
\begin{center}
    \scriptsize
    \setlength\tabcolsep{.9pt}
    \extrarowsep=1.5ex
    {\begin{tabular}{@{\extracolsep{.5 pt}} l c cc cc c @{}}
    \Xhline{2\arrayrulewidth}
      & \multicolumn{6}{c}{\bfseries PigArtPressure Set } \\ 
    \cline{2-7}
                           \multicolumn{1}{l}{\textbf{Option}}     
                        &  \multicolumn{1}{c}{\textbf{KNNDTW}} 
                        &  \multicolumn{1}{c}{\textbf{HMM-S}}
                        &  \multicolumn{1}{c}{\textbf{B-HMM-T}}
                        &  \multicolumn{1}{c}{\textbf{B-HMM-S}}
                        &  \multicolumn{1}{c}{\textbf{IOHMM}}
                        %&  \multicolumn{1}{c}{\textcolor{blue}{IOHMM}}                    
                        &  \multicolumn{1}{c}{\textbf{OUR}} \\                                  
    \Xhline{2\arrayrulewidth}
    N
    & .58    
    & .57(.05)
    & .40
    & .58
    & .44
    & .62(.05)   \\
    \hline
    RM25-FO
    & .61(.02) 
    & .56(.01)
    & .40
    & .58
    & .40(.03)
    & .61(.06)    \\
    
    RM25-NFO
    & \textemdash 
    & \textemdash
    & \textemdash
    & \textemdash    
    & \textemdash
    & .58(.04)    \\
    
    RM50-FO
    & .60(.02) 
    & .57(.03)
    & .40(.01)    
    & .58(.04)
    &  42(.04)
    & .59(.04)    \\

    RM50-NFO
    & \textemdash 
    & \textemdash
    & \textemdash    
    & \textemdash
    & \textemdash
    & .61(.05)    \\

    \hline
    
    IM25-FO
    & .43(.03) 
    & .46(.07)
    & .37(.04)
    & .52(.05)
    & .35(.04)
    & .62(.07)    \\
    
    IM25-NFO
    & \textemdash 
    & \textemdash
    & \textemdash
    & \textemdash
    & \textemdash
    & .59(.05)    \\    

    IM50-FO
    & .36(.01) 
    & .40(.05)
    & .31(.03)
    & .41(.03)
    & .34(.03)  
    & .57(.03)    \\
    
    IM50-NFO
    & \textemdash 
    & \textemdash
    & \textemdash
    & \textemdash
    & \textemdash
    & .59(.07)    \\
    
    \Xhline{2\arrayrulewidth}    
    \end{tabular}}    
    \caption{Estimation Result for PigArtPressure dataset including missing values (Q2)} 
\end{center}
\end{table}

% \section{\textcolor{blue}{Conclusion}}

\section{Conclusion}
We propose the scalable learning method for HMM-GPSM by combining SVI for a long sequence of state transition and R-RFF for a large number of time-series for GP emission. In experiments, we validate that the proposed learning method reduces learning time and estimates the state even though the dataset includes missing values.

In future research, we consider two research directions. First, we will extend the kernel's expressive power from stationary to non-stationary to overcome the limitation in modeling real data with a stationary kernel \cite{remes2017non,remes2018neural,ton2018spatial}. Second, we will relax the Markovian assumption over the transition of the hidden state by considering the complex sequence model such as RNN to figure out the complicated sequential relationship \cite{dai2016recurrent,linderman2016recurrent,linderman2017bayesian}.

\newpage

% \subsubsection*{Acknowledgements}

%%%%%%%%%%%%%%%%%%%%%%%%%%%%%%%%%%%%%%%%%%%%%%%%%%%%%%%%%%%%%%%%%%%%%%%%%%%%%%%%%%%%%%%%
%%%%%%%%%%%%%%%%%%%%%%%%%%%%%%%%%%%%%%%%%%%%%%%%%%%%%%%%%%%%%%%%%%%%%%%%%%%%%%%%%%%%%%%%
%%%%%%%%%%%%%%%%%%%%%%%%%%%%%%%%%%%%%%%%%%%%%%%%%%%%%%%%%%%%%%%%%%%%%%%%%%%%%%%%%%%%%%%%
%%%%%%%%%%%%%%%%%%%%%%%%%%%%%%%%%%%%%%%%%%%%%%%%%%%%%%%%%%%%%%%%%%%%%%%%%%%%%%%%%%%%%%%%
%%%%%%%%%%%%%%%%%%%%%%%%%%%%%%%%%%%%%%%%%%%%%%%%%%%%%%%%%%%%%%%%%%%%%%%%%%%%%%%%%%%%%%%%
%%%%%%%%%%%%%%%%%%%%%%%%%%%%%%%%%%%%%%%%%%%%%%%%%%%%%%%%%%%%%%%%%%%%%%%%%%%%%%%%%%%%%%%%

% \newpage

\bibliographystyle{unsrt}
\bibliography{bib}

% %%%%%%%%%%%%%%%%%%%%%%%%%%%%%%%%%%%%%%%%%%%%%%%%%%%%%%%%%%%%%%%%%%%%%%%%%%%%%%%%%%%%%%%%
% %%%%%%%%%%%%%%%%%%%%%%%%%%%%%%%%%%%%%%%%%%%%%%%%%%%%%%%%%%%%%%%%%%%%%%%%%%%%%%%%%%%%%%%%
% %%%%%%%%%%%%%%%%%%%%%%%%%%%%%%%%%%%%%%%%%%%%%%%%%%%%%%%%%%%%%%%%%%%%%%%%%%%%%%%%%%%%%%%%
% %%%%%%%%%%%%%%%%%%%%%%%%%%%%%%%%%%%%%%%%%%%%%%%%%%%%%%%%%%%%%%%%%%%%%%%%%%%%%%%%%%%%%%%%
% %%%%%%%%%%%%%%%%%%%%%%%%%%%%%%%%%%%%%%%%%%%%%%%%%%%%%%%%%%%%%%%%%%%%%%%%%%%%%%%%%%%%%%%%
% %%%%%%%%%%%%%%%%%%%%%%%%%%%%%%%%%%%%%%%%%%%%%%%%%%%%%%%%%%%%%%%%%%%%%%%%%%%%%%%%%%%%%%%%

% % References follow the acknowledgements.  Use an unnumbered third level
% % heading for the references section.  Any choice of citation style is
% % acceptable as long as you are consistent.  Please use the same font
% % size for references as for the body of the paper---remember that
% % references do not count against your page length total.

\newpage
\onecolumn
% \section{Supplementary Material}

\section{Supplementary Material}

\subsection{Derivation}
\subsubsection{SVI application to Bayesian HMM with GP emission}
We derive (1) how the ELBO of the full sequence of time series data is linearly approximated by the ELBO of the batch sequence of time series data and (2) how the batch factors introduced in Section 3.1 are derived. 

Given the sampled $i \in \{1,...,T-L+1\}$ uniformly, let $Y_{L}^{s} = \{y_{i,:},...,y_{i+L-1,:} \}$ be sampled observation with the length $L$ and $X_{L}^{s}$ be corresponding inputs and $Z_{L}^{s}$ corresponding hidden states. The expected log joint likelihood of $\{ Z_{L}^{s},Y_{L}^{s} \}$ given $X_{L}^{s}$ is approximated as 
\begin{align*}
&  \mathbb{E}_{s} \Big[ \mathbb{E}_{q} [\log{p(Y_{L}^{s},Z_{L}^{s} | X_{L}^{s} )}] \Big] \nonumber  \\
& = \sum_{i=0}^{T-L} \frac{1}{T-L+1}  \mathbb{E}_{q} [\log{p(Y_{L}^{s_{i}},Z_{L}^{s_{i}} | X_{L}^{s_{i}} )}]   \\
& = \frac{1}{T-L+1}  \sum_{i=0}^{T-L} 
 \mathbb{E}_{q} \left[  \log{p(z_{i})} +  
 \underbrace{\sum_{t=1}^{L} \log{p(A_{z_{i+t-1},z_{i+t}})}}_{\mathrm{transition \ term}}
+ \underbrace{\sum_{t=1}^{L} \log{p(y_{i+t,:}|z_{i+t},x_{i+t,:})}  }_{\mathrm{observation \ term}}
    \right]   \  \cdots \ (*)  \\
& \approx \frac{1}{T-L+1} \mathbb{E}_{q} \Big[ \sum_{t=1}^{T-L+1} \log{p(z_{t-1})} +  L \sum_{t=1}^{T} \log{p(A_{z_{t-1},z_{t}})} +  L \sum_{t=1}^{T} \log{p(y_{t,:}|z_{t},x_{t,:})} \Big]
\end{align*}
This implies that transition and observation term of $\mathbb{E}_{q} [\log{p(Y_{L}^{s},Z_{L}^{s} | X_{L}^{s} )}]$ for the sampled $i \in \{1,..T-L+1\}$ can be approximated as
\begin{align*}
\mathbb{E}_{q} \left[ \sum_{t=1}^{L} \log{p(A_{z_{i+t-1},z_{i+t}})}
\right] &\approx  \frac{L}{T-L+1}\mathbb{E}_{q} \left[ \sum_{t=1}^{T} \log{p(A_{z_{t-1},z_{t}})} \right] \\
\mathbb{E}_{q} \Bigg[\sum_{t=1}^{L} \log{p(y_{i+t,:}|z_{i+t},x_{i+t,:})}
\Bigg] &\approx  \frac{L}{T-L+1} \mathbb{E}_{q} \left[ \sum_{t=1}^{T} \log{p(y_{t,:}|z_{t},x_{t,:})}   \right]
\end{align*}
Thus, the batch factors, $c_{s}^{A}$ and $c_{s}^{\theta}$, to calibrate the approximated ELBO are obtained as 
\begin{align*}
c_{s}^{A} = \frac{T-L+1}{L} \ \ ,\ \ c_{s}^{\theta} = \frac{T-L+1}{L}
\end{align*}

The transition term in expectation in $(*)$ can be approximated as
\begin{align*}
\sum_{i=0}^{T-L} 
 \mathbb{E}_{q} \left[ \sum_{t=1}^{L} \log{p(A_{z_{i+t-1},z_{i+t}})} \right]  
 &= \mathbb{E}_{q} \left[ \sum_{j=1}^{L} \log{p(A_{z_{j-1},z_{j}})} + \sum_{j=2}^{L+1} \log{p(A_{z_{j-1},z_{j}})} + \cdots + \sum_{j=T-L+1}^{T} \log{p(A_{z_{j-1},z_{j}})}
  \right] \\
&= \mathbb{E}_{q} \left[ L \sum_{t=L}^{T-L+1} \log{p(A_{z_{t-1},z_{t}})} + 
\underbrace{\sum_{t=1}^{L-1} t \left(\log{p(A_{z_{t-1},z_{t}})} + \log{p(A_{z_{T-t},z_{T-t+1}})} \right)}_{\mathrm{approximated \ term}}
 \right] \\
&\approx \mathbb{E}_{q} \left[ L \sum_{t=L}^{T-L+1} \log{p(A_{z_{t-1},z_{t}})} + L \sum_{t=1}^{L-1}  \left(\log{p(A_{z_{t-1},z_{t}})} + \log{p(A_{z_{T-t},z_{T-t+1}})} \right)\right] \\
&= \mathbb{E}_{q} \left[L \sum_{t=1}^{T} \log{p(A_{z_{t-1},z_{t}})} \right] \\
\end{align*}
Here, the observation term in expectation in $(*)$ can be approximated as
\begin{align*}
\sum_{i=0}^{T-L} 
 &\mathbb{E}_{q} \left[ \sum_{t=1}^{L} \log{p(y_{i+t,:}|z_{i+t},x_{i+t,:})} \right] =
  \mathbb{E}_{q} \left[ \sum_{j=1}^{L} \log{p(y_{j,:}|z_{j},x_{j,:})} + \cdots + \sum_{j=T-L+1}^{T} \log{p(y_{j,:}|z_{j},x_{j,:})} \right] \\
&= \mathbb{E}_{q} \Bigg[ L \sum_{t=L}^{T-L+1} \log{p(y_{t,:}|z_{t},x_{t,:})} +
\underbrace{ \sum_{t=1}^{L-1} t \left( \log{p(y_{t,:}|z_{t},x_{t,:})} + \log{p(y_{T-t+1,:}|z_{T-t+1},x_{T-t+1,:})} \right) }_{\mathrm{approximated \ term}} \Bigg]\\
&\approx \mathbb{E}_{q} \Bigg[ L \sum_{t=L}^{T-L+1} \log{p(y_{t,:}|z_{t},x_{t,:})} + L \sum_{t=1}^{L-1}  \left( \log{p(y_{t,:}|z_{t},x_{t,:})} + \log{p(y_{T-t+1,:}|z_{T-t+1},x_{T-t+1,:})} \right)\Bigg]\\
&= \mathbb{E}_{q} \left[ L \sum_{t=1}^{T} \log{p(y_{t,:}|z_{t},x_{t,:}) } \right]
\end{align*}

\subsubsection{Reparameterized RFF SM kernel approximation}
 Given the parameters of SM kernel $\{w_{q},{\mu}_{q},\sigma_{q}\}_{q=1}^{Q}$, we sample spectral points from Gaussian distribution $p^{(q)}(S) = N(S;{\mu}_{q},\sigma_{q})$ by reparametrization trick. 
\begin{align*}
& s_{q,i} = \mu_{q} + \sigma_{q} \circ \epsilon_{i} 
% \ \ \forall \ i=1,..,m_q
\end{align*}
where $\epsilon_{i} \sim N(\epsilon;0,I)$ for $i=1,..,m_q$. 

If we define the $x_{S^{(q)}} = [2 \pi x^{T} s_{q,1},.., 2 \pi x^{T} s_{q,m_{q}}]$ and the feature map $\phi_{q}(x) = \frac{1}{\sqrt m_q } [\cos( x_{S^{(q)}}), \sin(x_{S^{(q)})}]$, then $\phi_{q}(x)\phi_{q}(y)^{T}$  can approximate $k_q(x-y)$ that is the inducted kernel from Gaussian Spectral density $p^{(q)}(S)$ by Bochner's theorem.
\begin{align*}
\mathbb{E}_{p^{(q)}(S)} \left[ \phi_{q}(x)\phi_{q}(y)^{T} \right]  &= \mathbb{E}_{p^{(q)}(S)} 
\Bigg[
\frac{1}{m_q} \sum_{i=1}^{m_q} \left( \cos{{  2\pi s_{(q,i)}}^{T}x} \right) \left( \cos{{2\pi s_{(q,i)}}^{T}y} \right)+ \left( \sin{{  2\pi s_{(q,i)}}^{T}x} \right) \left( \sin{{2\pi s_{(q,i)}}^{T}y} \right)
\Bigg]
\\
&= \mathbb{E}_{p^{(q)}(S)} 
\Bigg[
\frac{1}{m_q} \sum_{i=1}^{m_q} \cos{ {2\pi s_{(q,i)}}^{T}(x-y)} 
\Bigg] \\
&= \mathbb{E}_{p^{(q)}(S)} 
\Bigg[
\frac{1}{m_q} \sum_{i=1}^{m_q} \frac{ e^{ i {2\pi s_{(q,i)}}^{T}(x-y) } + e^{-i {2\pi s_{(q,i)}}^{T}(x-y) } }{2}  
\Bigg] \\
&= \frac{1}{2} \left( k_{q}(x-y) + k_{q}(y-x) \right) = 
\frac{1}{2} \left( k_{q}(x-y) + k_{q}(x-y) \right)\\
&= k_{q}(x-y)
\end{align*}
Using the above derivation, we define sampled spectral points $\textbf{\em s} = \cup_{q=1}^{Q} \{ s_{q,i} \}_{i=1}^{m_q} $ from $p(S)=\prod_{q=1}^{Q} p^{(q)}(S)$ and the feature map $\phi_{SM}(x) = \left[ \sqrt{w_{1}}\phi_{1}(x),..,\sqrt{w_{Q}}\phi_{Q}(x) \right]$, which can approximate $k_{SM}(x,y)$ by $\phi_{SM}(x)\phi_{SM}(y)^{T}$.
\begin{align*}
\mathbb{E}_{p(S)} \left[ \phi_{SM}(x)\phi_{SM}(y)^{T} \right]  &=
\mathbb{E}_{p(S)} \left[ \sum_{q=1}^{Q}w_{q}\phi_{q}(x)\phi_{q}(y)^{T} \right] \\
&=\sum_{q=1}^{Q}w_{q} \mathbb{E}_{p^{(q)}(S)} \left[ \phi_{q}(x)\phi_{q}(y)^{T} \right]\\
&=\sum_{q=1}^{Q}w_{q} k_q(x-y) = k_{SM}(x-y)
% &  k_{SM}(x-y) \approx \phi_{SM}(x)\phi_{SM}(y)^{T}
\end{align*}

\subsubsection{Regularized Lower bound for GP emission}
We consider the lower bound of log marginal likelihood with the candidate distribution $q(S)$. We can derive the lower bound $\mathcal{L}$ as follows:
\begin{align*}
\log{p(Y|X)} &= \log{ \iint p(Y|f)p(f|X,S)\frac{p(S)}{q(S)}q(S) df d S } \\  
&= \log{ \int p(Y|X,S)\frac{p(S)}{q(S)}q(S) d S } \\  
&\geq \int \log { \Big( p(Y|X,S)\frac{p(S)}{q(S)} \Big) } q(S)  d S \\
&= \int \log { p(Y|X,S)q(S) } + \log { \frac{p(S)}{q(S)} } q(S)  d S  \\
&= \int \log { p(Y|X,S)q(S) }d S - KL(q(S)||P(S))    =  \mathcal{L} \\ 
\end{align*}
Applying the Stochastic Gradient Variational Bayes (SGVB) \cite{kingma2013auto} to $\mathcal{L}$ with the reparametrizable distribution $q(S)$, leads to the following unbiased estimator $\hat{\mathcal{L}_{K}}$. 
\begin{align*}
\hat{\mathcal{L}_{K}} &= \frac{1}{K}\sum_{i=1}^{K}\log { p(Y|X,s^{(i)}) } - KL(q(S)||P(S))
\end{align*}
where $s^{(i)}$ is $i$-th sampled spectral points from $q(S)$.

%%%%%%%%%%%%%%%%%%%%%%%%%%%%%%%%%%%%%%%%%%%%%%%%%%%%%%%%%%%%%%%%%%%%%%%%%%%%%%%%%%%
%%%%%%%%%%%%%%%%%%%%%%%%%%%%%%%%%%%%%%%%%%%%%%%%%%%%%%%%%%%%%%%%%%%%%%%%%%%%%%%%%%%
\newpage

\subsection{Algorithm}

\raggedbottom

\subsubsection{Initialization of SM Kernel hyperparameters} 
% Spectral mixture kernel possess many hyperparameters. For effective training, initialization of SM kernel hyperparameters is very important. We take the following initialization scheme to consider the empirical spectral density.

\centering{
    \begin{minipage}{0.8\textwidth}
    \begin{algorithm}[H]
        \SetAlgoLined
        \LinesNumberedHidden
        \DontPrintSemicolon
        \SetKwInOut{Input}{Input}
        \SetKwInOut{Output}{Output}
        \Input{Data : X = $\{x_{t,:}\}_{t=1}^{T}$ and Y = $\{y_{t,:}\}_{t=1}^{T}$ 
        \\ K : \#Cluster \ , \  $Q$ : \#Mixture Component for SM kernel}
        \Output{$\theta = \{ {\{w^{k}_{q},{\mu}^{k}_{q},\sigma^{k}_{q}\}} \}_{q=1,k=1}^{Q,K}$}
        \For{t=1,..,T}{
            Get the normalized empirical spectral density $s_{t}$ from $y_{t}$ by FFT \\
            }
        Get the mean of $\{m_{k}\}_{k=1}^{K}$ for each cluster after applying k-means clustering $s_{t}$\\
        
        \For{k=1,..,K}{
            Get the cumulative distribution function (CDF) of $m_{k}$ by applying CumSum to $m_{k}$\\
            Sample the spectral points by applying inverse sampling the CDF of $m_{k}$ \\
            Fit the Gaussian Mixture distribution (GMM) by the sampled spectral points \\
            Obtain $k$ th SM kernel parameters $\{w^{k}_{q},{\mu}^{k}_{q},\sigma^{k}_{q}\}_{q=1}^{Q}$ 
            }

        Get the initial SM hyperparameters 
        $\{w^{k}_{q},{\mu}^{k}_{q},\sigma^{k}_{q}\}_{q=1,k=1}^{Q,K}$
            
        \caption{SM Kernel parameters initialization for Clustering }
        \end{algorithm}
    \end{minipage}
}

\subsubsection{GPSM(RSS)-SVI } 
\centering{    
    \begin{minipage}{0.8\textwidth}
        \begin{algorithm}[H]
        \SetAlgoLined
        \LinesNumberedHidden
        \DontPrintSemicolon      
        \SetKwInOut{Input}{Input}
        \SetKwInOut{Output}{Output}
        \Input{Data : X = $\{x_{t,:}\}_{t=1}^{T}$ and Y = $\{y_{t,:}\}_{t=1}^{T}$, 
        \\ $K$ : \#Cluster \ , \ $Q$ : \#Mixture Component for SM kernel,
        \\ $L$ : the length of sampled sequence \ , \ $M$ : \#batch for sampled sequence
        \\ $m$ : \#Spectral point for SM kernel approximation
        }
        \Output{
            $\theta^{*} = \{ {\{w^{k}_{q},{\mu}^{k}_{q},\sigma^{k}_{q}\}} \}_{q=1,k=1}^{Q,K}$, $w^{*}_{\pi}$, and $w^{*}_{A}$
        }
        % \LinesNotNumbered
        Initialize SM kernel hyperparameter $\theta$ by \textbf{Algorithm 1} \\
        \While{ not converged }{
            Sample the $m$ batch data $\{Y^{s_m}_{L},X^{s_m}_{L}\}_{m=1}^{M}$    \\
            \# Local Variable Parameter Update \\
            \For{$m = 1:M$}{
                Calculate $q^{*}(Z^{s_m}_{L})$ for $ \{Y^{s_m}_{L},X^{s_m}_{L}\} $ by with equation $(17)$ and $(18)$  \\
            }
           \# Global Variable Parameter Update \\
            Update $w_{pi}$ and $w_{A}$ by equation $(27)$ and $(28)$ \\
            Update SM kernel hyperparmeters $\theta $ by applying ADAM to the equation (29) \\
            except that $\log(p(y_{i+t,:},z_{i+t},y_{i+t,:})$ in (29) is replaced by the equation (32) 
        }
        \caption{GPSM(RSS)-SVI Learning}         
        \end{algorithm} 
    \end{minipage}
}

%%%%%%%%%%%%%%%%%%%%%%%%%%%%%%%%%%%%%%%%%%%%%%%%%%%%%%%%%%%%%%%%%%%%%%%%%%%%%%%%%%%
%%%%%%%%%%%%%%%%%%%%%%%%%%%%%%%%%%%%%%%%%%%%%%%%%%%%%%%%%%%%%%%%%%%%%%%%%%%%%%%%%%%

%%%%%%%%%%%%%%%%%%%%%%%%%%%%%%%%%%%%%%%%%%%%%%%%%%%%%%%%%%%%%%%%%%%%%%%%%%%%%%%%%%%
%%%%%%%%%%%%%%%%%%%%%%%%%%%%%%%%%%%%%%%%%%%%%%%%%%%%%%%%%%%%%%%%%%%%%%%%%%%%%%%%%%%
\newpage

\subsection{Experiments}

\subsubsection{Scalable Learning (Q1)}

\begin{figure}[ht]
\centering{
    \subfloat[GPSM-SVI Loss]{
    \includegraphics[width=0.23\textwidth]{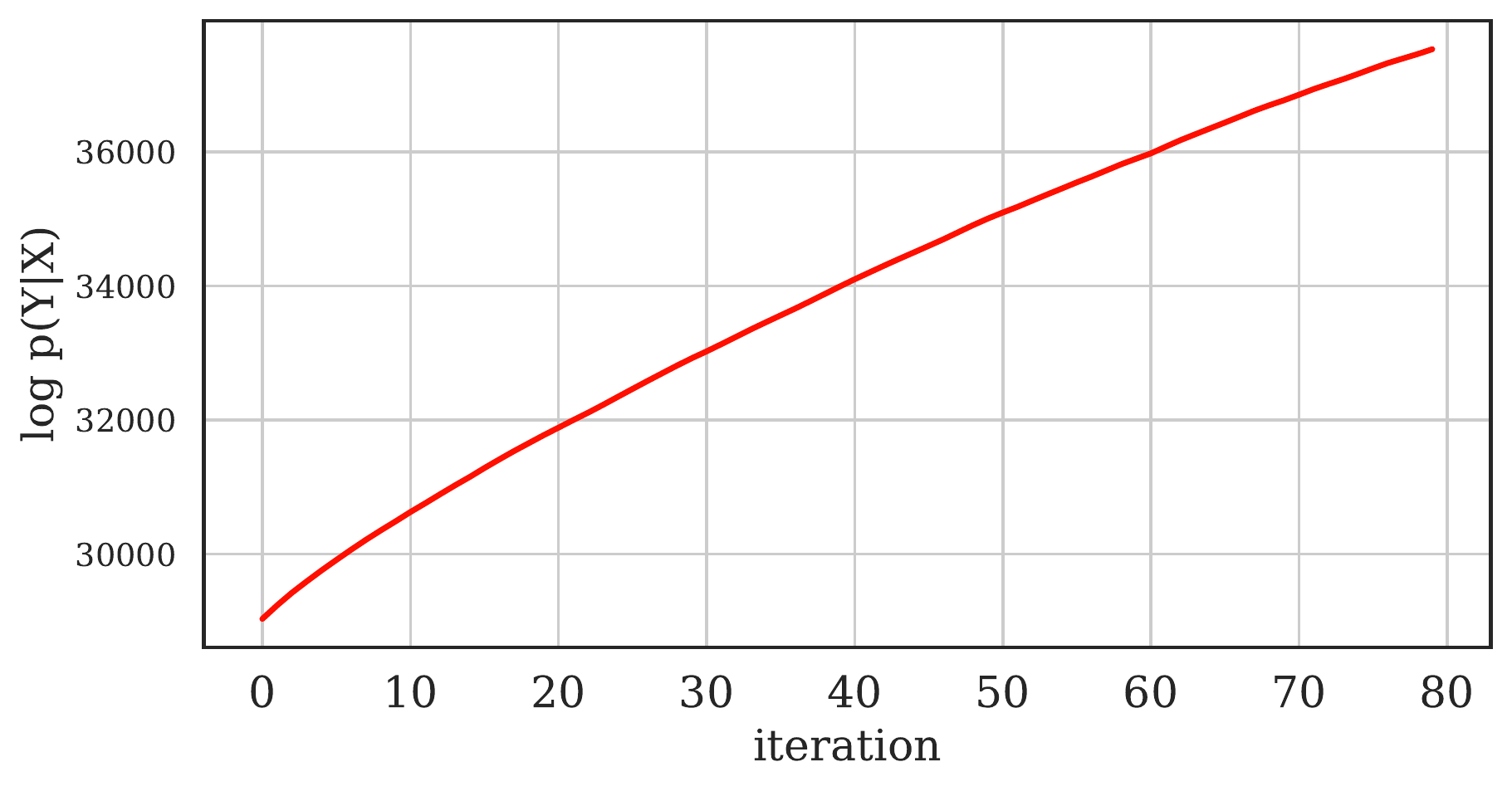}}
    \hfill
    \subfloat[GPSM-SVI Accuracy]{
    \includegraphics[width=0.23\textwidth]{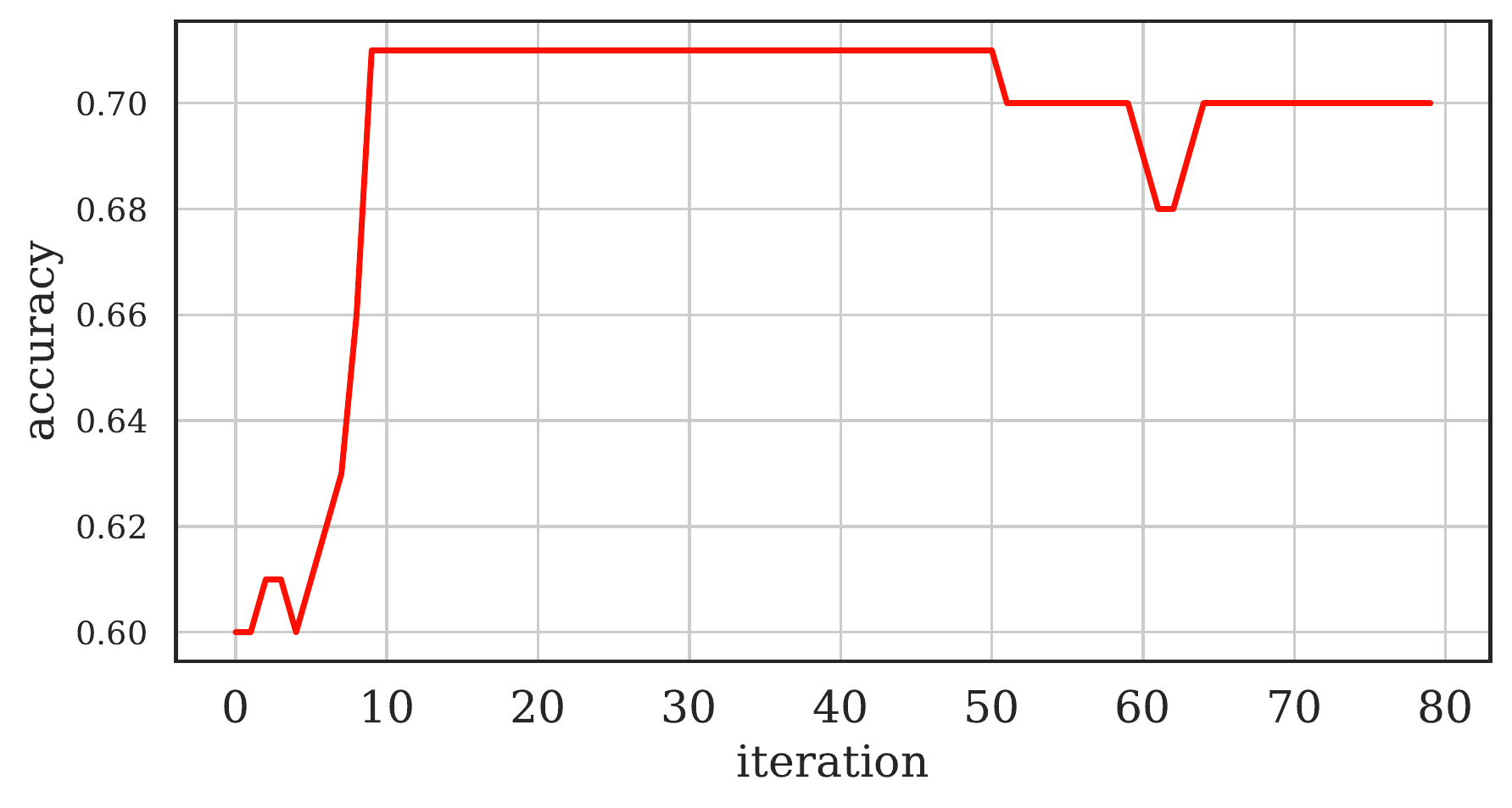}}
    \hfill    
    \subfloat[GPSM(RSS)-SVI Loss]{    
    \includegraphics[width=0.23\textwidth]{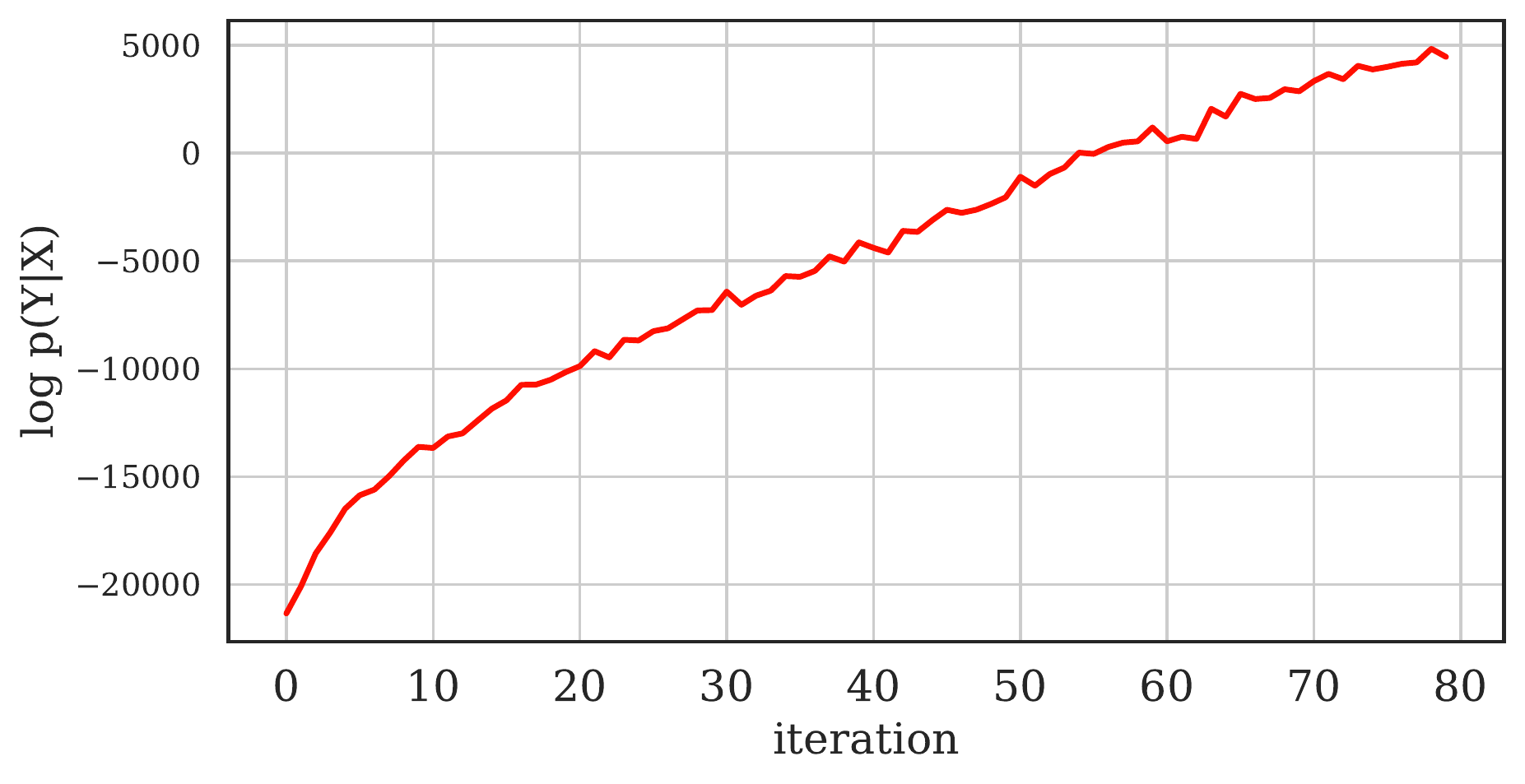}}
    \hfill    
    \subfloat[GPSM(RSS)-SVI Accuracy]{
    \includegraphics[width=0.23\textwidth]{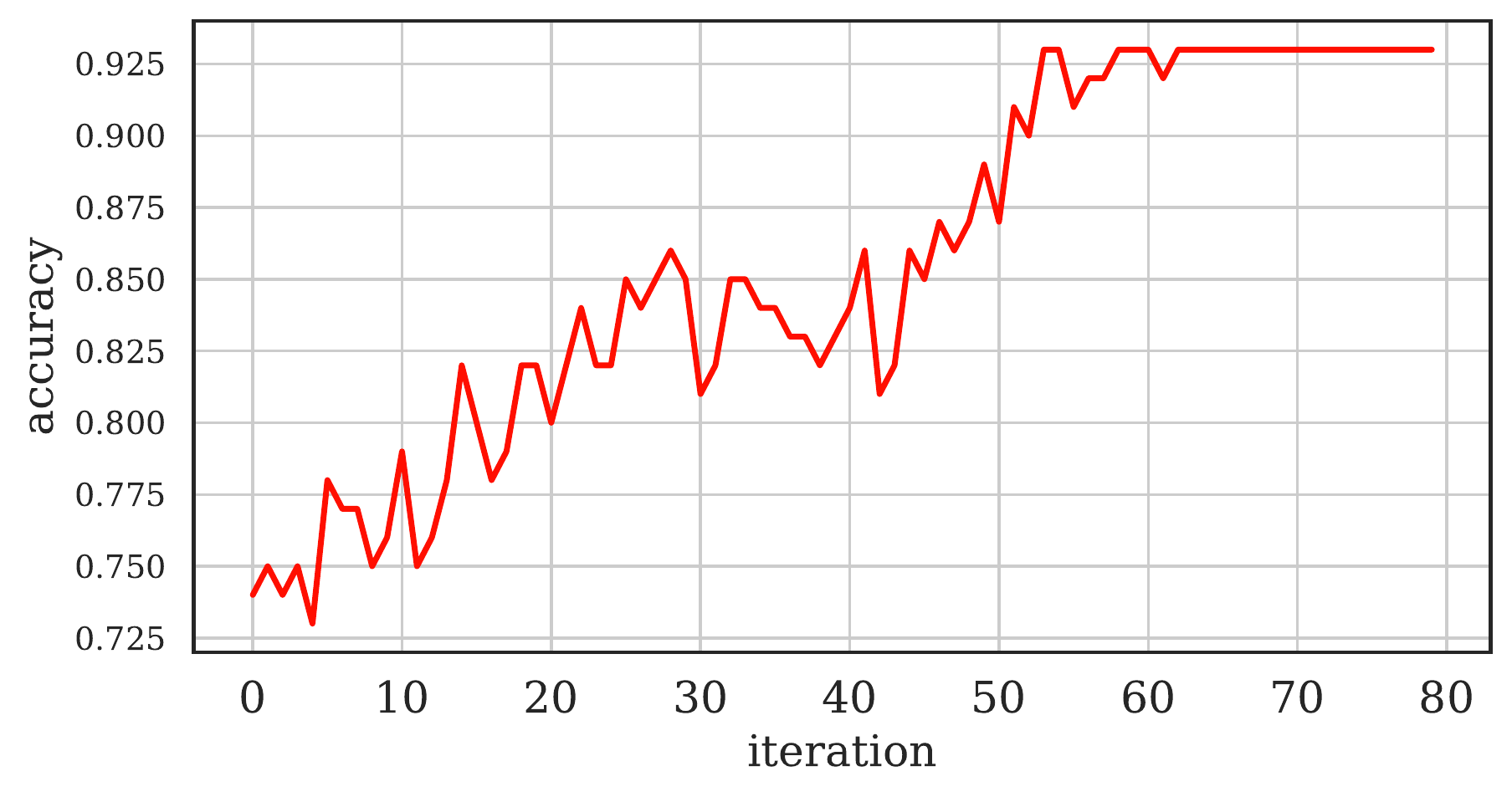}}
    \hfill    
    \newline
    \subfloat[Q=6,Hz=1000 (True)  with Accuracy 1.0 ]
    {
    \includegraphics[width=.95\textwidth,height=1.7cm]{Supplementary_EXP1/_dataset_Q6_1000Hz.pdf}
    }
    \hfill
    \subfloat[GPSM-SVI with Accuracy .73]
    {
    \includegraphics[width=.95\textwidth,height=1.7cm]{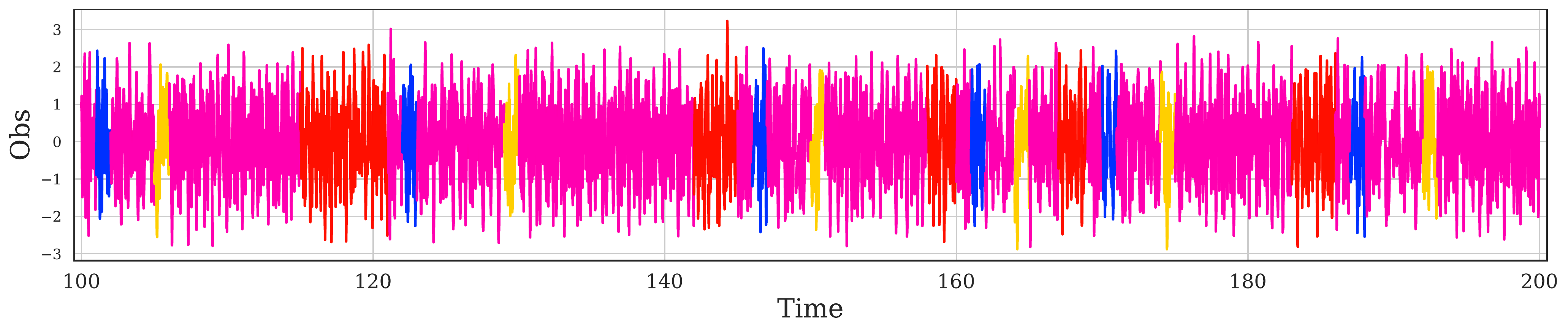}
    }
    \hfill
    \subfloat[GPSM(RSS)-SVI with Accuracy .94]
    {
    \includegraphics[width=.95\textwidth,height=1.7cm]{Supplementary_EXP1/_dataset_Q6_1000Hz_ssgpr_reg_SVI_10_3_4_10_Acc94.pdf}
    }
    \hfill    
    \subfloat[BHMM-S with Accuracy .78]
    {
    \includegraphics[width=.95\textwidth,height=1.7cm]{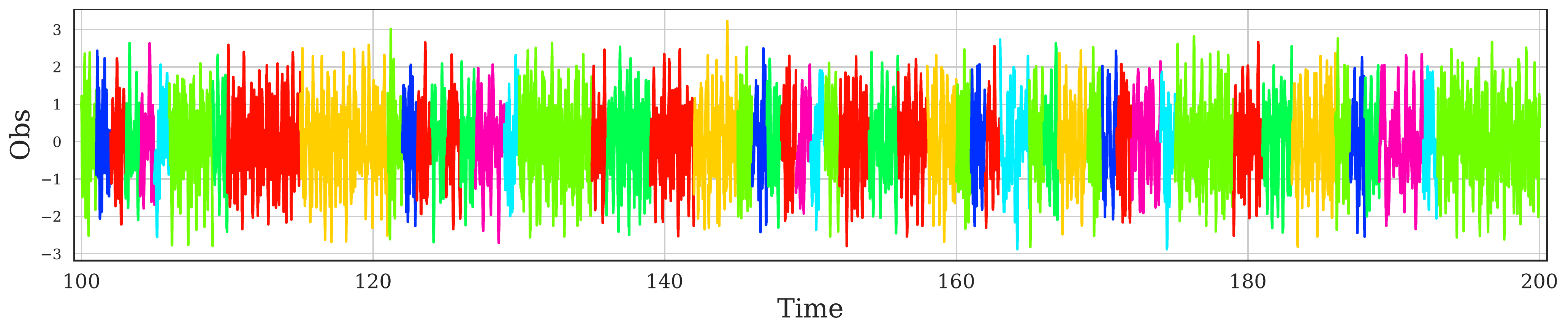}
    }
    \hfill    
    \subfloat[KNNDTW with Accuracy .68]
    {
    \includegraphics[width=.95\textwidth,height=1.7cm]{Supplementary_EXP1/_dataset_Q6_1000Hz_KNNDTW.pdf}
    }
}
\caption{We describe the scalable learning experiment result for the case of $Q=6$, $Hz=1000$. Our models are trained during 80 iterations until the models arrive at local optimal. Figures (a) and (c) show the log marginal likelihood $\log{p(Y|X)}$ of GPSM-SVI and GPSM(RSS)-SVI during training. Figures (b) and (d) show the change of the training Accuracy during training.  Figures (e), (f), (g), (h), and (i) shows each model's clustering result for test data along with Accuracy.}    
\end{figure}

\subsubsection{Robust State Estimation (Q2)}

\begin{figure}[ht]
\centering{
    \subfloat[PigArtPressure Testset]
    {
    \includegraphics[width=.95\textwidth,height = 1.4cm]{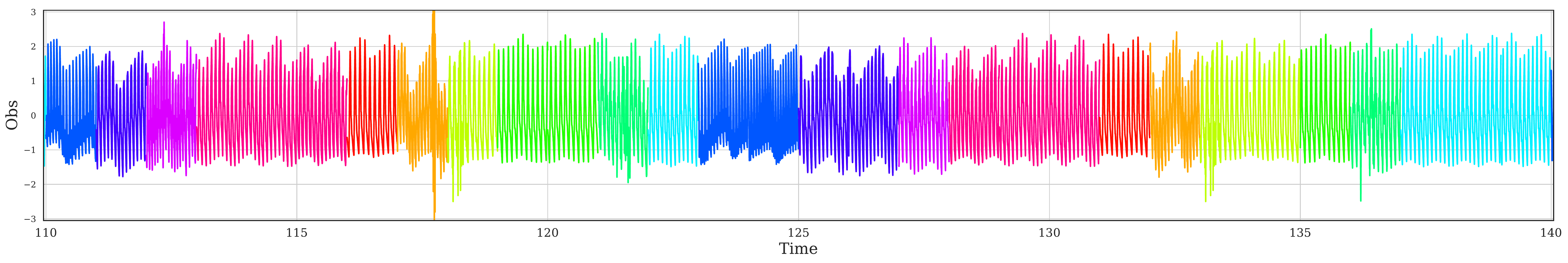}}
    \hfill
    
    \subfloat[RM50-FO : BHMM-S  with Accuracy .58 ]
    {
    \includegraphics[width=.95\textwidth,height = 1.4cm]{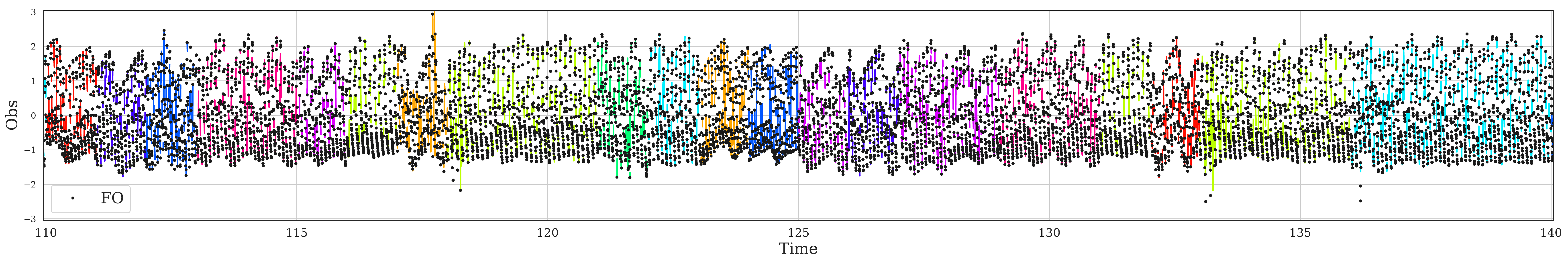}}
    \hfill
    
    \subfloat[RM50-FO : IOHMM with Accuracy .48 ]
    {
    \includegraphics[width=.95\textwidth,height = 1.4cm]{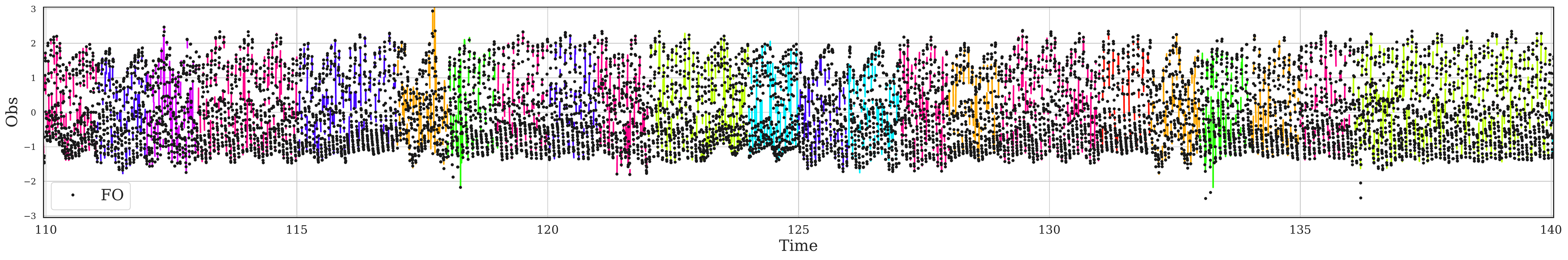}}    
    \hfill
    
    \subfloat[RM50-FO : GPSM(RSS)-SVI  with Accuracy .62]
    {
    \includegraphics[width=.95\textwidth,height = 1.4cm]{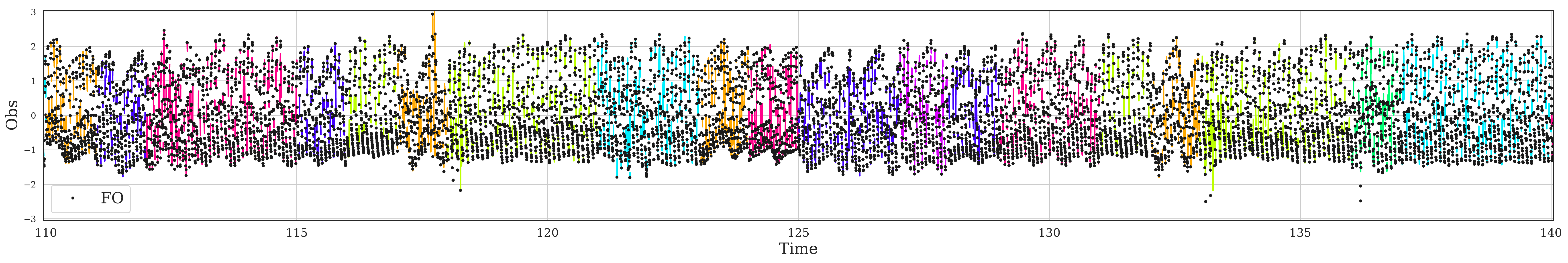}}    
    \hfill
    
    \subfloat[RM50-NFO : GPSM(RSS)-SVI  with Accuracy .66]
    {
    \includegraphics[width=.95\textwidth,height = 1.4cm]{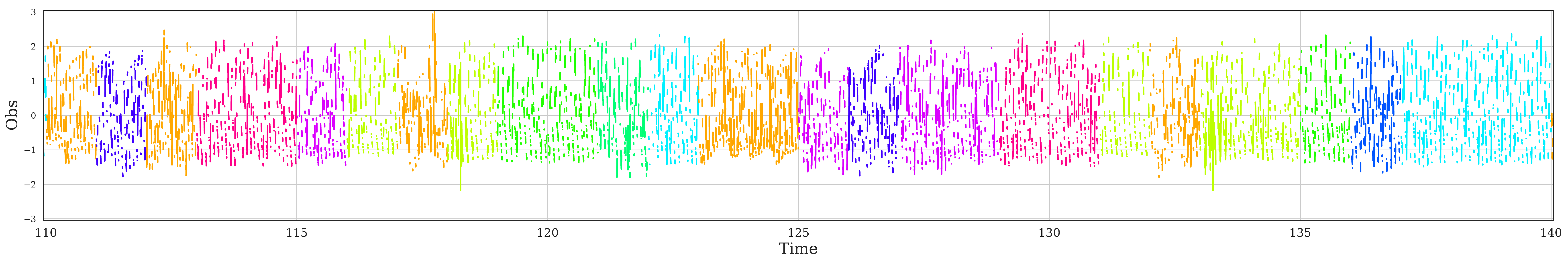}}    
    \hfill
    
    % %%%%%%%%%%%%%%%%%%%%%%%%%%%%%%%%%%%%%%%%%%%%%%%%%%%%%%%%%%%%%%%%%%%%%%%%
    % %%%%%%%%%%%%%%%%%%%%%%%%%%%%%%%%%%%%%%%%%%%%%%%%%%%%%%%%%%%%%%%%%%%%%%%%

    \subfloat[IM25-FO : BHMM-S  with Accuracy .50 ]
    {
    \includegraphics[width=.95\textwidth,height = 1.4cm]{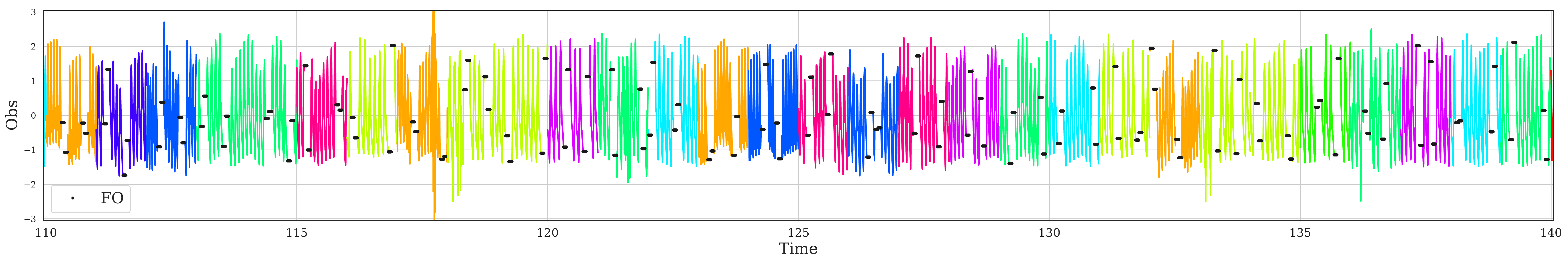}}
    \hfill

    \subfloat[IM25-FO : IOHMM with Accuracy .32 ]
    {
    \includegraphics[width=.95\textwidth,height = 1.4cm]{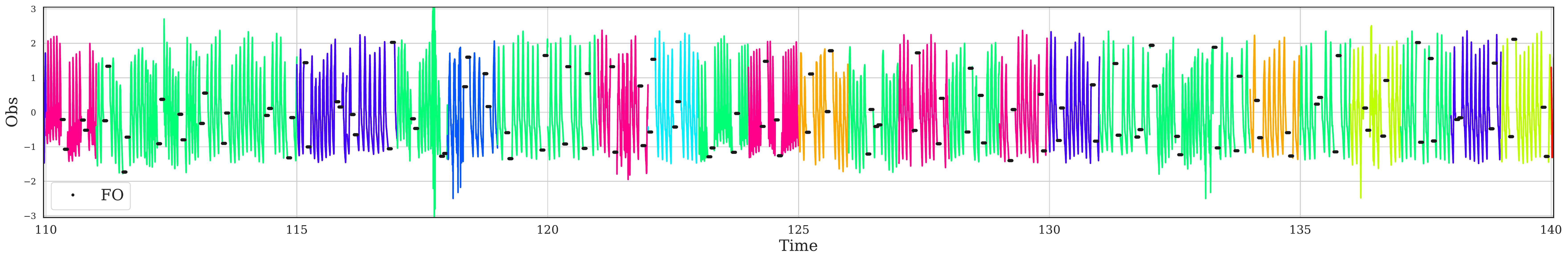}}    
    \hfill
    
    \subfloat[IM25-FO : GPSM(RSS)-SVI  with Accuracy .72 ]
    {
    \includegraphics[width=.95\textwidth,height = 1.4cm]{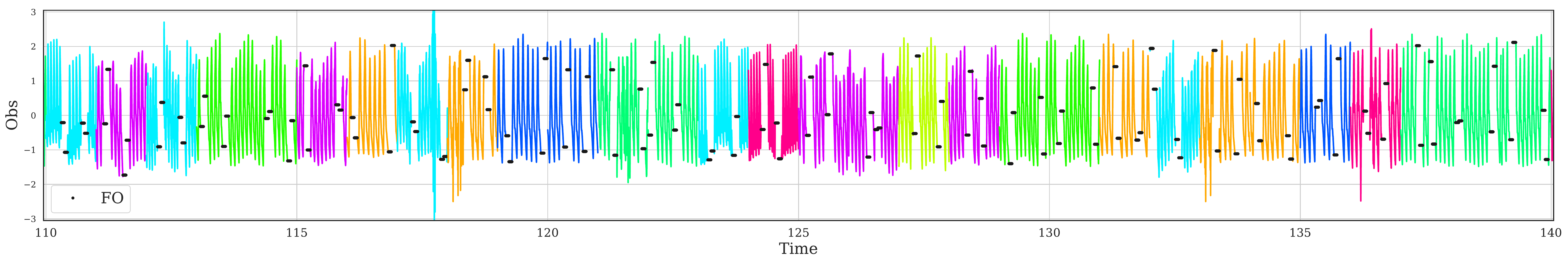}}
    \hfill
    
    \subfloat[IM25-NFO : GPSM(RSS)-SVI  with Accuracy 0.70 ]
    {
    \includegraphics[width=.95\textwidth,height = 1.4cm]{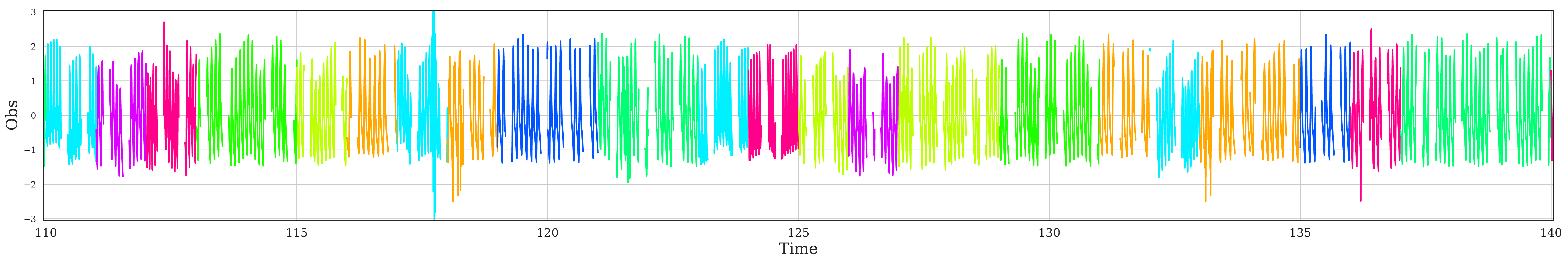}}
}
\caption{We describe the state estimation result for experiment 2 (Robust estate estimation). Figure (a) shows a subset of test data within the time interval [110,140] out of test data set [100,150]. Figures (b),(c), and (d) show the results of the clustering task for the case of RM50 with the FO technique. Figure (e) shows the clustering results for GPSM(RSS)-SVI (our proposed method) without using FO. Similarly, Figures (f),(g), and (h) show the results for IM25. Figure (i) shows the clustering result for GPSM(RSS)-SVI (our proposed method) without using the FO technique for IM25.}
\end{figure}

\end{document}